\documentclass[sn-apa,iicol]{sn-jnl}

\usepackage{times}
\usepackage{epsfig}
\usepackage{graphicx}
\usepackage{comment}
\usepackage{amsmath,amssymb} 
\usepackage{color}
\usepackage{subfigure}
\usepackage{booktabs}
\usepackage{verbatim}
\usepackage{algorithmicx,algorithm}
\usepackage{graphicx}
\usepackage{amsmath}
\usepackage{amssymb}
\usepackage{booktabs}
\usepackage{xcolor}
\usepackage{multirow}
\usepackage{makecell}
\usepackage{pifont}
\usepackage{url}

\newcommand{\etal}{\textit{et al.}}
\newcommand{\eg}{\textit{e.g.}}
\newcommand{\ie}{\textit{i.e.}}

\jyear{2021}%

\theoremstyle{thmstyleone}%
\usepackage{marvosym}
\theoremstyle{thmstyletwo}%
\theoremstyle{thmstylethree}%
\begin{document}
	
	\title[VFM4AD]{Forging Vision Foundation Models for Autonomous Driving:  Challenges, Methodologies, and Opportunities}
	
	
	\author[1]{Xu Yan$^*$}
	\author[2,1]{Haiming Zhang$^*$}
	\author[1]{Yingjie Cai$^*$}
	\author[1]{Jingming Guo$^*$}
	\author[1]{Weichao Qiu$^*$}
	\author[1]{Bin Gao$^*$}
	\author[1]{Kaiqiang Zhou$^*$}
	\author[1]{Yue Zhao$^*$}
	\author[1]{Huan Jin$^*$}
	\author[1]{Jiantao Gao$^*$}
	\author[2]{Zhen Li}
	\author[1]{Lihui Jiang}
	\author[1]{Wei Zhang}
	\author[1]{Hongbo Zhang}
 	\author[3]{Dengxin Dai}
	\author[1]{Bingbing Liu$^{\textrm{\Letter}}$}

	\affil[1]{Huawei Noah's Ark Lab}
	\affil[2]{SSE, The Chinese University of Hong Kong (Shenzhen)}
 	\affil[3]{Huawei Zurich Research Center} 
	\affil{\textit{\small $^*$Equal Contributions}}

	\def\ie{\textit{i.e.}}
	\def\eg{\textit{e.g.}}
	\def\etal{\textit{et al.}}
	\def\etc{\textit{etc}}

	\abstract{
		The rise of large foundation models, trained on extensive datasets, is revolutionizing the field of AI. Models such as SAM, DALL-E2, and GPT-4 showcase their adaptability by extracting intricate patterns and performing effectively across diverse tasks, thereby serving as potent building blocks for a wide range of AI applications.
		Autonomous driving, a vibrant front in AI applications, remains challenged by the lack of dedicated vision foundation models (VFMs). 
        The scarcity of comprehensive training data, the need for multi-sensor integration, and the diverse task-specific architectures pose significant obstacles to the development of VFMs in this field.
	This paper delves into the critical challenge of forging VFMs tailored specifically for autonomous driving, while also outlining future directions. Through a systematic analysis of over 250 papers, we dissect essential techniques for VFM development, including data preparation, pre-training strategies, and downstream task adaptation. Moreover, we explore key advancements such as NeRF, diffusion models, 3D Gaussian Splatting and world models, presenting a comprehensive roadmap for future research.
		To empower researchers, we have built and maintained \href{Forge_VFM4AD}{https://github.com/zhanghm1995/Forge_VFM4AD}, an open-access repository constantly updated with the latest advancements in forging VFMs for autonomous driving.
	}

	\keywords{Vision Foundation Models, Data Generation, Self-supervised Training, Autonomous Driving,  Literature Survey.}
	
	
	
	\maketitle

	\section{Introduction}
	\label{sec:intro}
	
	The rapid progress in autonomous driving (AD) technology is reshaping the transportation landscape, ushering in an AI-driven future.
	Traditional autonomous driving perception systems rely on a modular architecture, utilizing dedicated algorithms for specific tasks, such as object detection~\cite{lang2019pointpillars,mao2021voxel}, semantic segmentation~\cite{guo2018review,yan20222dpass}, and depth estimation~\cite{ming2021deep}.
	Each task is typically addressed by a separate model, typically a deep neural network trained on task-specific labels.
	However, these compartmentalized components prioritize individual task performance at the expense of broader contextual understanding and data relationships. This approach often results in output inconsistencies and limits the system's ability to handle long-tail cases.
	
	\begin{figure*}[t]
		\begin{center}
			\includegraphics[width=\linewidth]{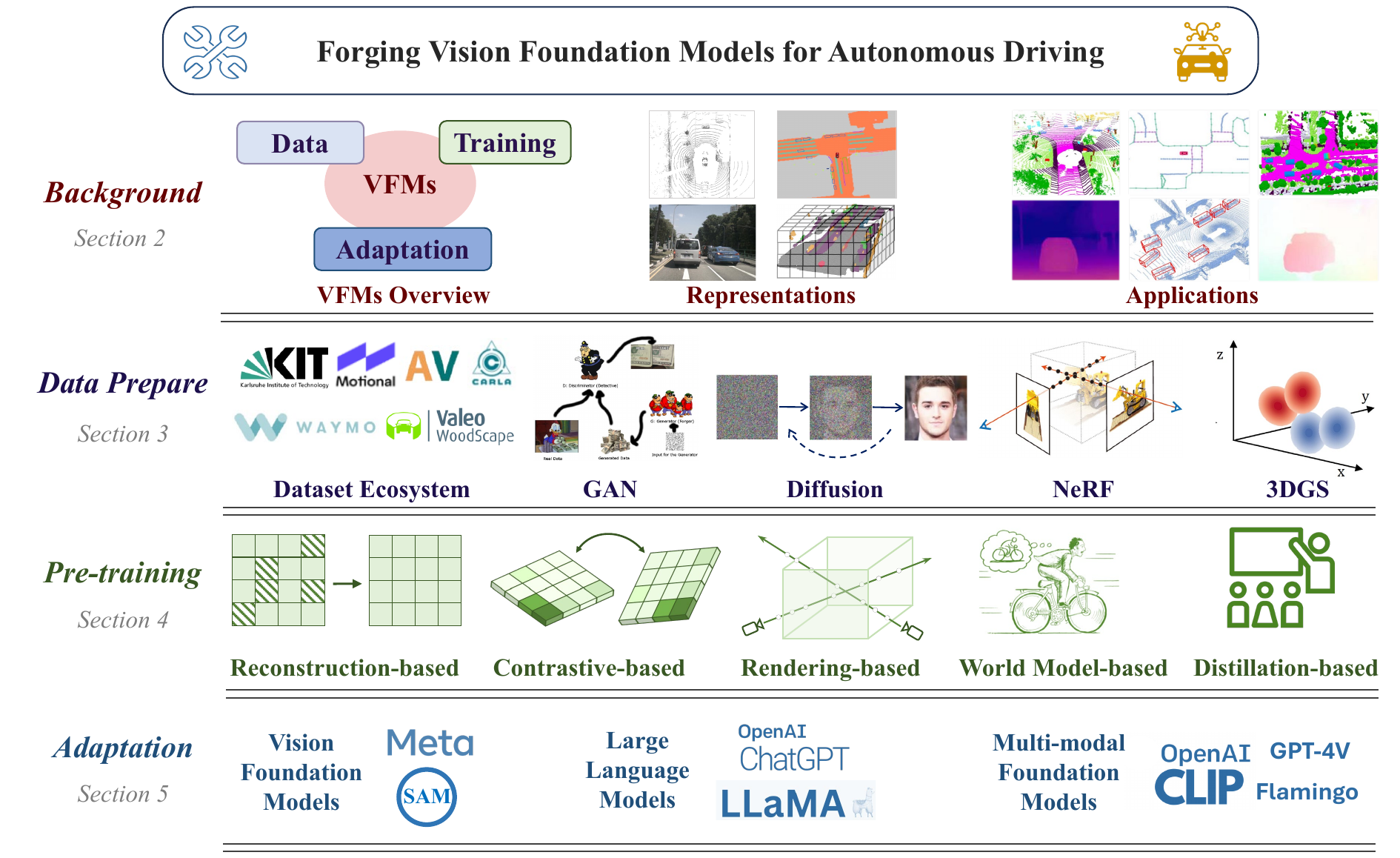}
		\end{center}
		
		\caption{\textbf{Our survey at a glance.} \textbf{Background.} Tis section first introduces the development of foundation models, while also delving into the diverse representations and applications within the autonomous driving community. 
			\textbf{Data Preparation.} The challenge of amassing substantial volumes of data for training foundational models is particularly pronounced in the context of vision-based foundational models for autonomous driving. Our investigation encompasses an in-depth analysis of existing autonomous driving datasets acquired through on-road collection and simulators, as well as noteworthy advancements such as generative adversarial networks (GAN), diffusion models, neural radiance field (NeRF), and 3D Gaussian Splatting (3DGS) techniques.
			\textbf{Pre-training.} Self-supervised learning constitutes a pivotal aspect of our exploration. We categorize prevalent self-supervised pre-training methods into reconstruction-based, contrastive-based, distillation-based, rendering-based, and world model-based approaches. 
			\textbf{Adaptation.} In bridging the gap between trained Vision Foundation Models (VFMs) and downstream tasks, we investigate the application of VFMs developed in other domains to the autonomous driving field. We acknowledge the use of images from online resources and published papers.}
		\label{fig:overview}
		
	\end{figure*}
	
	Large-scale foundation models, particularly those in Natural Language Processing (NLP)~\cite{openai2023gpt4,brown2020language}, have emerged as a powerful force in the field of artificial intelligence. 
	These models, trained on vast and diverse datasets, often leverage self-supervision techniques. Once trained, they can be adapted through fine-tuning to tackle a wide range of specific tasks with one model.
	The recent success of billion-parameter models like GPT-3/4~\cite{brown2020language,openai2023gpt4} in zero/few-shot learning~\cite{ateia2023chatgpt,espejel2023gpt,liang2023breaking} is particularly noteworthy. 
	Their remarkable ability for few-shot learning positions them well to effectively handle scenarios with out-of-distribution AD data, such as encountering unforeseen objects. Furthermore, their inherent capacity for reasoning makes them highly suitable for tasks requiring logical processing and informed decision-making.
	
	While large foundation models have indeed revolutionized various domains, their impact on AD has not met expectations.
	Directly applying existing vision foundation models (VFMs) trained on 2D data or text modalities from other domains to AD tasks has proven to be demonstrably insufficient. 
	These models lack the capacity to leverage the rich 3D information crucial for AD perception tasks, such as depth estimation.
	Moreover, the intrinsic heterogeneity of AD architectures and the necessity for multi-sensor fusion present additional challenges to the direct adaptation of VFMs.
	This challenge is further compounded by the imperative for a VFM capable of efficiently processing diverse sensor data (\eg, LiDAR, camera, radar) and seamlessly adapting to various downstream tasks within the AD domain.

	In the context of autonomous driving development, two key factors hinder the progress of vision foundation models:
	\begin{itemize}
		\setlength{\itemsep}{0pt}
		\setlength{\parsep}{0pt}
		\setlength{\parskip}{0pt}
		\item[-] \textbf{Data Scarcity:}  AD data is inherently limited due to privacy concerns, safety regulations, and the complexity of capturing real-world driving scenarios. Moreover, AD data must meet strict requirements, including multi-sensor alignment (\eg, LiDAR, cameras, radars), and temporal consistency.
		\item[-] \textbf{Task Heterogeneity:} 
        Autonomous driving presents a range of diverse tasks, each requiring distinct input forms (\eg, camera, LiDAR, radar) and output formats (\eg, 3D bounding boxes, lane lines, depth maps). This heterogeneity poses a challenge for VFMs, as architectures optimized for one task often perform unsatisfactory on others. Consequently, developing a single, general-purpose architecture and representation that efficiently handles multi-sensor data and performs well across disparate downstream tasks remains a significant obstacle.

	\end{itemize}
	
	Despite these challenges, there are promising indications that the development of large vision foundation models for self-driving is on the horizon. The increasing availability of data through continuous collection~\cite{caesar2020nuscenes,mao2021one} and the development of advanced simulation technologies~\cite{yang2023unisim,li2023drivingdiffusion} provide the potential to address the issue of data scarcity. 
	Additionally, the recent advancements in perception, particularly the shift towards a unified representation utilizing bird's-eye view (BEV)~\cite{philion2020lift,li2022bevformer}, and occupancy~\cite{tian2023occ3d}, offer a potential solution to the problem of lacking a generalizable representation and architecture.

	This paper delves into the key technologies underpinning the development of large vision foundation models for autonomous driving, as shown in Fig.~\ref{fig:overview}. 
	Our exploration begins by establishing a comprehensive background on foundation models, existing frameworks, and tasks, as well as the development of representation, outlining our core motivations in Sec.\ref{background}.
	Subsequently, we delve into existing datasets and data simulation techniques in Sec.~\ref{data}, highlighting the crucial role of technologies like Generative Adversarial Networks (GANs), Neural Radiance Fields (NeRFs), Diffusion Models, and 3D Gaussian Splatting (3DGS) in addressing the inherent data scarcity in autonomous driving. 
	Building upon this foundation, Sec.~\ref{training} analyzes available self-training techniques for efficiently training VFMs on unlabeled real-world data. 
	Finally, to bridge the gap between trained VFMs and downstream tasks, Sec.~\ref{adapt} explores the application of foundation models developed in other domains to the AD field. We examine the valuable lessons learned and potential adaptations for achieving effective performance across diverse downstream tasks within autonomous driving.

	In contrast to existing survey papers~\cite{yang2023survey,huang2023applications,firoozi2023foundation,sun2023survey} encompassing the application of large foundation models across various domains, this paper presents a novel approach by focusing on the development of large vision foundation models tailored specifically to address the challenges of autonomous driving.
	This unique perspective enables us to delve deeper into the fundamental principles and technological advancements essential for constructing VFMs capable of driving substantial progress in the field.
	
\begin{figure*}[t]
		\begin{center}
			\includegraphics[width=\linewidth]{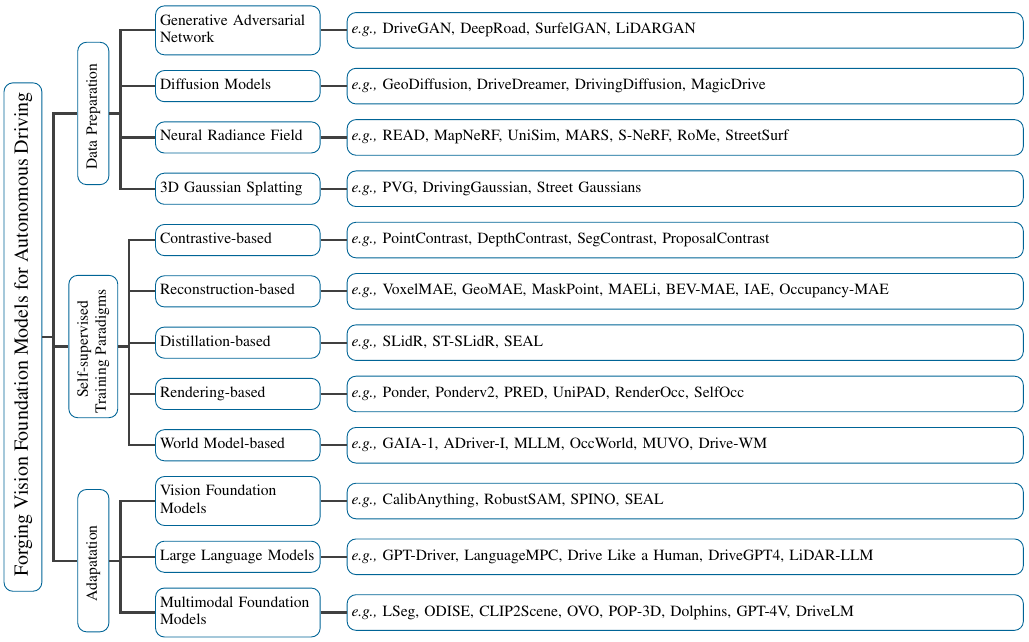}
		\end{center}
		
		\caption{\textbf{Research tree of forging vision foundation models for autonomous driving.}}
		\label{fig:mind}
	\end{figure*}

	The major contributions of this work can be summarized as follows:
	\begin{itemize}
		\setlength{\itemsep}{0pt}
		\setlength{\parsep}{0pt}
		\setlength{\parskip}{0pt}
		\item[-] We adopt a unifying pipeline for developing large vision foundation models (VFMs) for autonomous driving. This pipeline encompasses comprehensive reviews of data preparation, self-supervised learning, and adaptation.
		\item[-] We systematically categorize existing works across each process within the proposed framework, as shown in Fig.~\ref{fig:mind}. Our analysis offers fine-grained classifications, in-depth comparisons, and summarized insights in each section.
		\item[-] We delve into the critical challenges encountered in forging VFMs for autonomous driving. Drawing insights from over 250 surveyed papers, we summarize key aspects and propose future research directions.
	\end{itemize}

    \section{Background}
	\label{background}
	This section begins with the landscape of existing large foundation models, followed by the evolution of representation and common tasks in autonomous driving perception. 
	
	\subsection{Large Foundation Models}
	Foundation models are pre-trained on large-scale datasets.
	The landscape of large foundation models is rapidly evolving, with diverse architectures pushing the boundaries of machine intelligence. 
	
	\noindent\textbf{Large Language Models.} 
	The emergence of large language models (LLMs) such as GPT-3~\cite{brown2020language} has brought about a revolution in natural language processing. These AI powerhouses train from vast amounts of text data, enabling them to comprehend and generate language with astonishingly human-like fluency. From text completion and translation to dialogue and question answering, LLMs showcase expertise across a wide range of linguistic tasks.
	
	Driven by advancements in research and training methods, a diverse array of even more sophisticated LLMs has surfaced. Notable examples include GPT-4~\cite{openai2023gpt4}, the driving force behind ChatGPT, and PaLM~\cite{chowdhery2022palm}, the core of Bard. Additionally, open-source options like LLaMA/LLaMA2~\cite{touvron2023llama,touvron2023llama2} have gained traction, offering parameter counts ranging from 7 billion to a substantial 65 billion.
	Multilingual support has also become a prominent focus, with models such as ChatGLM-6B~\cite{du2022glm} and GLM-130B~\cite{zeng2022glm} showcasing their multilingual capabilities and fine-tuning potential, thereby expanding their reach and adaptability.
	
	\noindent\textbf{Vision Foundation Models.} 
	Inspired by the success of large language models, the field of computer vision has also embraced similarly potent models. 
	Vision Transformers (ViTs)~\cite{dosovitskiy2020image} tackle visual tasks with comparable power. Their transformer architecture, trained on massive datasets like LVD-142M~\cite{oquab2023dinov2}, excels in tasks like image classification and segmentation. DINO~\cite{caron2021emerging} and DINOv2~\cite{oquab2023dinov2} further enhance these capabilities by leveraging self-supervised learning with knowledge distillation. Features learned using DINO contain explicit information about the scene layout and object boundaries.
	Other methods like MAE~\cite{caron2021emerging}, BEIT~\cite{caron2021emerging}, and CAE~\cite{caron2021emerging} employ masked modeling as a powerful self-supervised learning technique, enabling them to learn general visual representations. 
	
	More recently, the Segment Anything Model (SAM)~\cite{kirillov2023segment} has emerged as a virtuoso in object segmentation, adeptly crafting precise masks for individual elements within images. Its training on a vast dataset encompassing 11 million images and 1.1 billion masks has endowed it with the exceptional ability to generalize to diverse segmentation tasks without any explicit fine-tuning, showcasing remarkable zero-shot performance.

		

    \begin{figure*}[t]
		\begin{center}
			\includegraphics[width=\linewidth]{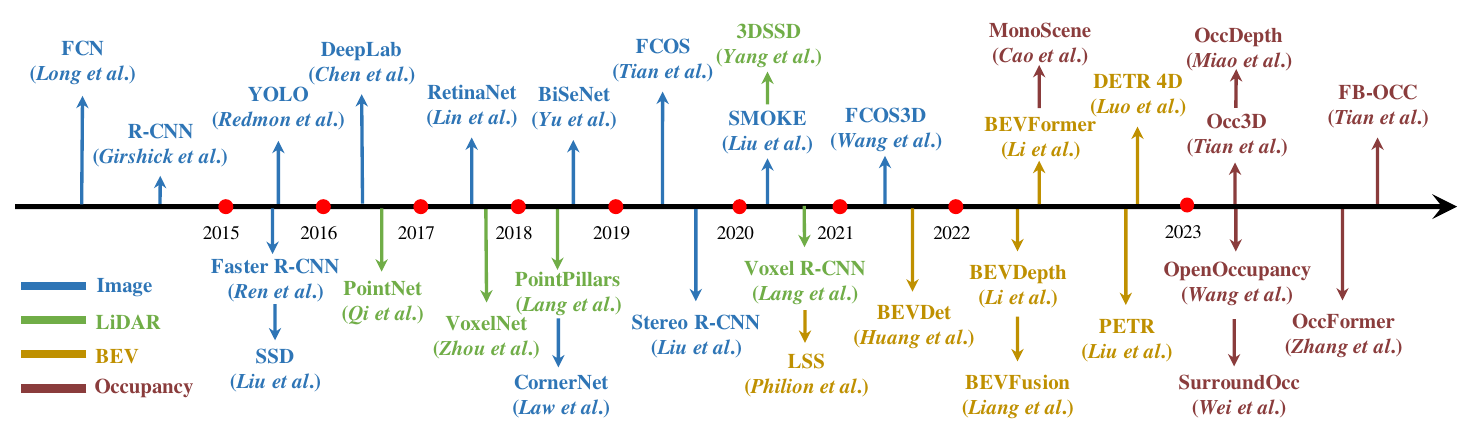}
		\end{center}
		
		\caption{\textbf{Chronological overview of the image, LiDAR, BEV and occupancy representations.} Only representative approaches are demonstrated.}
		\label{fig:framework}
	\end{figure*}
 
	\noindent\textbf{Multimodal Foundation Models.}  
	In the realm of computer vision, Multimodal Vision-Language Models (VLMs) such as CLIP~\cite{radford2021learning} and BLIP~\cite{li2022blip} serve as a crucial bridge between text and vision. Trained on vast image-text pairs from the internet, they adeptly capture the intricate relationships between visual and textual information. This enables them to excel in tasks like zero-shot image classification and instance-level alignment (\eg, CLIP2~\cite{zeng2023clip2} and FILIP~\cite{yao2021filip}) with remarkable accuracy.
	
	The field of computer vision increasingly explores the power of combining diverse pre-trained foundation models.
	For instance, SAMText~\cite{SAMText} empowers precise text segmentation by automatically generating pixel-level masks around detected text, leveraging information from pre-existing detectors.
	Similarly, Caption Anything~\cite{wang2023caption} establishes a versatile framework for image captioning, enabling interactive manipulation of both visual and textual aspects. By synergistically merging SAM and ChatGPT, users can dynamically refine images through various prompts, including pointing or drawing bounding boxes directly.
	Moreover, GPT-4V(ision)~\cite{openai2023gpt4} unlocks a deeper understanding and analysis of user-provided image inputs, showcasing the evolution of multimodal capabilities in computer vision.

	
	\noindent\textbf{Generative Foundation Models.} 
	Visual Generative Models like DALL-E and DALL-E2~\cite{reddy2021dall} push the boundaries of creativity. Trained on extensive datasets of text-image pairs, these models excel in generating stunning images from scratch based on textual descriptions.
	The recent emergence of Stable Diffusion~\cite{rombach2022high} has sparked a wave of creativity in combining its diffusion-based image generation with existing methods. One prime example is Inpaint Anything~\cite{yu2023inpaint}, which seamlessly integrates LaMa~\cite{suvorov2022resolution} and Stable Diffusion for inpainting masked regions. By leveraging text prompts and these powerful models, users can seamlessly generate specific content to fill or replace voids within an image. Similarly, Edit Everything~\cite{gao2023editanything} showcases a versatile generative system that harnesses the combined strengths of SAM, CLIP, and Stable Diffusion. This fusion empowers users to manipulate images with stunning precision, guided by both visual cues and textual prompts.


    
	\subsection{Development of Representation}
 
	This section provides a comprehensive overview of the key representations utilized in autonomous driving perception, encompassing images, point clouds, bird's-eye view (BEV), and occupancy grids. Some typical approaches are demonstrated in Fig.~\ref{fig:framework}.
	
	\noindent\textbf{Image as Representation.} 
	Leveraging the rich texture information inherent in RGB images, monocular cameras are adopted in perception tasks in primary autonomous driving. This inherent advantage naturally fosters a straightforward 2D output approach, where these images serve as a foundational representation and dedicated networks are designed accordingly~\cite{girshick2014rich,ren2016faster,redmon2016look,lin2018focal,law2019cornernet}.
	However, the inherent limitation of monocular cameras lies in their inability to directly perceive depth information. To achieve 3D results, they necessitate transforming the 2D output to the 3D space.
	The following works~\cite{tian2019fcos,wang2021fcos3d,liu2020smoke} address this challenge by first estimating 2D locations, orientations, and dimensions from extracted features.  Subsequently, they undertake the 3D task by transforming these intermediate estimations.
	Recognizing this limitation, researchers have increasingly turned to stereo camera systems, capitalizing on the synergy of spatial and temporal cues to enrich 3D perception tasks~\cite{pon2020objectcentric,li2019stereo,qin2019triangulation}.

	\noindent\textbf{Point Cloud as Representation.}  LiDAR sensors have emerged as a cornerstone in autonomous driving, thanks to their exceptional depth sensing capabilities and rich 3D geometric information, exceeding the abilities of conventional cameras.
	
	Several pioneering studies have established point clouds as the foundational representation for LiDAR-based perception tasks. Among these approaches, four main streams prevail: point-based, voxel-based, projection-based and hybrid-based approaches.
	Point-based methods~\cite{qi2017pointnet,yan2020pointasnl,hu2019randla} prioritize the original geometry of raw point clouds, leveraging permutation-invariant operators to capture intricate local structures and fine-grained patterns without compromising data fidelity through quantization.
	Voxel-based methods~\cite{zhou2017voxelnet,lang2019pointpillars,deng2021voxel,malashin2021sparsely} excel at transforming irregular point clouds into compact 3D grids, facilitating efficient processing and integration with traditional convolutional neural networks.
	Projection-based methods~\cite{wu2019squeezesegv2, liong2020amvnet,kong2023rethinking} adopt a highly efficient approach, projecting point clouds onto 2D pixels (\textit{e.g.,} range images) to leverage the powerful capabilities of established 2D-CNN architectures.
	Hybridizing these diverse approaches presents further intriguing possibilities. Hybrid-based methods~\cite{tang2020searching,shi2020pv,xu2021rpvnet} seek to synergistically combine the strengths of different representation schemes, potentially opening doors to novel and even more accurate perception models.
	
	Furthermore, point clouds offer a valuable intermediate representation for facilitating multi-sensor fusion in object detection and scene understanding. Notably, works like~\cite{vora2020pointpainting} demonstrate this potential by projecting 2D features onto point clouds via ``feature painting", thereby enriching the perceptual capabilities of the perception.

	\noindent\textbf{Bird's-Eye View (BEV).} 
	BEV representation offers inherent advantages for autonomous driving perception tasks. Unlike the perspective view, BEV provides an unobstructed top-down view, eliminating occlusion issues and mitigating scale variations. This makes it highly amenable for subsequent modules like path planning and control, where object positions and relationships are crucial.
	
	LiDAR's native 3D structure makes it particularly adept at generating BEV representations. LiDAR-based methods readily collapse the height dimension of point cloud features to obtain a BEV map.
	Camera-based methods~\cite{ma2023visioncentric,li2023delving,li2023fastbev}, however, require additional steps to bridge the gap between 2D images and the 3D BEV space. Depth-based lifting~\cite{huang2022bevdet,li2022bevdepth} leverages depth estimation, while query-based lifting~\cite{li2022bevformer,liu2022petr,luo2022detr4d} utilizes specific queries to extract relevant features from the image and project them onto the BEV plane.
	Once a unified BEV feature representation is obtained, downstream tasks can be tackled through dedicated task-specific heads or by exploiting modality fusion techniques~\cite{liu2022bevfusion}.

	\noindent\textbf{Scene as Occupancy.} While BEV approaches have been predominant in the field of scene understanding, occupancy perception represents an emerging paradigm with unique advantages. In contrast to BEV's 2D projection, occupancy perception directly encodes the 3D environment using 3D voxel grids, enabling precise detection of obstacles in the crucial vertical dimension. This capability allows for robust handling of overhanging structures such as bridges, tunnels, and tree branches, which pose significant challenges for BEV methods. Furthermore, occupancy models can seamlessly integrate semantic information and velocity estimates, leading to a richer and more accurate representation of the surrounding scene.
	
	Pioneering works in this burgeoning domain introduce new benchmarks using nuScenes dataset~\cite{caesar2020nuscenes} at the same period, such as OpenOccupancy~\cite{wang2023openoccupancy}, OpenOcc~\cite{tong2023scene}, SurroundOcc~\cite{wei2023surroundocc} and Occ3D~\cite{tian2023occ3d}.
	These works mainly adopt the architecture from BEV perception and use 3D convolution to construct an extra head for occupancy prediction.
	Subsequent works with specific designs~\cite{gan2023simple,cao2022monoscene,miao2023occdepth,zhang2023occformer}  showcase the diverse approaches and promising potential of occupancy perception.

	\subsection{Vision Applications}
	%
	%
	A robust and reliable perception system requires the ability to understand the surrounding driving environment such as obstacles, traffic signs, and the free drivable areas in front of the vehicle.
	This section will delve into various downstream tasks that can refine and adapt pre-trained vision models to achieve this critical level of environmental understanding. 
	
	
	\noindent\textbf{Depth Estimation.} The objective of depth estimation is to produce dense depth maps from input images. These methods can be broadly categorized into two groups: stereo-based and monocular-based. Stereo methods~\cite{laga2020survey} utilize triangulation from two overlapping viewpoints, necessitating precise camera calibration. Conversely, monocular methods~\cite{eigen2014depth,alhashim2018high,lee2022self} estimate depth from a single image, garnering increased attention due to their simpler setup and reduced calibration requirements.

	\noindent\textbf{Object Detection.} Object detection aims to predict the locations, sizes, and classes of critical objects, \eg, cars, cyclists, and pedestrians.
	Generally, object detection can be categorized into 2D object detection~\cite{ren2015faster,jiang2022review} and 3D object detection~\cite{zhou2018voxelnet,shi2019pointrcnn}. The former exclusively acquires 2D bounding boxes on images, while the latter requires knowledge of the actual distance information of objects from the ego vehicles~\cite{mao20233d}.

	\noindent\textbf{Map Construction.} High-definition (HD) maps contain an enriched semantic understanding of elements on the road, serving as a fundamental module for navigation and path planning in autonomous driving. Traditional offline pipelines for building HD maps demand a substantial amount of human effort~\cite{kim2021hd,wang2022meta}. Recently, online semantic map construction~\cite{wang2023lidar2map,zhang2023online} has garnered increasing attention. Leveraging camera, LiDAR, or multi-sensor inputs, these approaches can generate rich information about the road layout.

	\noindent\textbf{Semantic Segmentation.}
	Semantic segmentation plays a pivotal role in autonomous driving and can be categorized into 2D semantic segmentation~\cite{chen2017deeplab,mohan2021efficientps} and 3D semantic segmentation~\cite{zhang2019review}. The former aims to assign semantic labels to each pixel, while the latter involves assigning semantic labels to each point in a 3D point cloud.

	\noindent\textbf{Object Tracking.} Object tracking~\cite{braso2022multi,guo2022review} aims to continuously estimate the position and movement of individual objects over time within the surrounding environment, assigning and maintaining unique IDs for each object to enable consistent tracking through changes in appearance and temporary occlusions.
	
	\noindent\textbf{Occupancy Prediction.} Recently, 3D occupancy prediction has garnered significant attention, aiming to jointly estimate the occupancy state and semantic label of every voxel in driving scenes from images~\cite{tian2023occ3d,li2023fbocc}. Compared with compact 3D bounding boxes, the occupancy representation excels in representing general objects, backgrounds, and irregularly shaped objects.

	\begin{table*}[t]
\centering
\caption{\textbf{Existing perception datasets for driving scene.} For \textbf{Supported Tasks}, \textit{OD}: Object Detection \textit{SS}: Semantic Segmentation \textit{OT}: Object Tracking \textit{MP}: Motion Planning. For \textbf{Data Diveristy}, \textit{Scenes} indicates the number of video clips. \textit{SU, SN, C, F, R} in \textit{Weather} indicate Sunny, Snowy, Cloudy, Foggy, and Rainy respectively. ``-" indicates that a field is inapplicable. Some statistics data are adopted from~\cite{li2023open}.}
\resizebox{\textwidth}{!}{
	\begin{tabular}{l|c|ccc|cccc}
		\toprule
		\multirow{2}{*}{\textbf{Dataset}} &  \multirow{2}{*}{\textbf{Supported Tasks}} & \multicolumn{3}{c|}{\textbf{Data Diversity}} &\multicolumn{4}{c}{\textbf{Sensors}} \\ ~ & ~& Scenes & Hours & Weather & Camera  & LiDAR & Radar &Others \\
		\hline
		
		Caltech Pedestrian~\cite{dollar2009pedestrian} &  2D OD & - & 10 & SU, C & Front-view & \ding{55} & \ding{55} & - \\
		KITTI \cite{geiger2012we} &  2D/3D OD, SS, OT & 50 & 6 & SU, C & Front-view & \ding{51} & \ding{55} & GPS, IMU \\
		Cityscapes~\cite{cordts2016cityscapes} &  2D/3D OD, 2D SS & - & - & SU, C & Front-view & \ding{55} & \ding{55} & - \\
		HDD \cite{ramanishka2018toward} &  2D OD & - & 104 & SU, C & Front-view & \ding{51} & \ding{55} & GPS, IMU, CAN \\
		IDD \cite{varma2019idd} &  2D SS & 182 &- & SU, C & Front-view & \ding{55} & \ding{55} & - \\
		DrivingStereo~\cite{yang2019drivingstereo} &  2D SS & 42 &- & SU, C, F, R & Front-view & \ding{51} & \ding{55} & GPS, IMU \\
		WoodScape~\cite{yogamani2019woodscape} &  2D/3D OD, 2D SS & - & - & SU, C & $360^{\circ}$ & \ding{51} & \ding{55} & GPS, IMU, CAN \\
		ApolloScape~\cite{huang2019apolloscape} &  2D/3D OD, 2D/3D SS, OT & 103 &2.5 & SU, C, R & Front-view & \ding{51} & \ding{55} & GPS, IMU  \\
		Brno-Urban~\cite{ligocki2020brno} &  2D/3D OD & 67 &10 & SU, C, R & Front-view & \ding{51} & \ding{55} & GPS, IMU, Infrared Camera  \\
		SemanticKITTI \cite{behley2019semantickitti} &  3D SS & 50 &6 & SU, C & \ding{55} & \ding{51} & \ding{55} & GPS, IMU \\
		Argoverse 1~\cite{Argoverse} &  3D DD, 3D SS, OT, MP & 113 & - & SU, C & $360^{\circ}$ & \ding{51} & \ding{55} & GPS \\
		KITTI-360~\cite{Liao2022PAMI} &  3D OD, 2D/3D SS & 366 & 2.5 & SU, C &$360^{\circ}$ & \ding{51} &\ding{55} & GPS, IMU \\ 
		nuScenes~\cite{caesar2020nuscenes} &  2D/3D OD, 2D/3D SS, OT, MP & 1,000 &5.5 & SU, R & $360^{\circ}$ & \ding{51} & \ding{51} & GPS, IMU, CAN \\
		BDD-100K~\cite{yu2020bdd100k} &  2D OD, 2D SS, OT & - & - & SU, C & Front-view & \ding{51} & \ding{55} &-  \\
		Waymo~\cite{sun2020scalability} &  2D/3D OD, 2D SS, OT, MP & 1,000 & 6.4 & SU, C, R& $360^{\circ}$ & \ding{51} & \ding{55} & GPS, IMU \\
		A2D2 \cite{geyer2020a2d2} &  3D OD, 2D SS & - & 5.6 & SU, C, R & $360^{\circ}$ & \ding{51} & \ding{55} & GPS, IMU, CAN \\
		SemanticPOSS~\cite{pan2020semanticposs} &  3D SS & - & - & SU, C & \ding{55} & \ding{51} & \ding{55} & GPS, IMU \\
		ONCE \cite{mao2021one} &  2D/3D OD, 2D SS, OT & - &144 & SU, C & $360^{\circ}$ & \ding{51} & \ding{55} & - \\
		Argoverse 2~\cite{wilson2022argoverse} &  3D OD, 3D OT & 1,000 & 4 & SU, C, SN, R & $360^{\circ}$ & \ding{51} & \ding{55} & GPS \\
		ZOD~\cite{alibeigi2023zenseact} &  2D/3D OD, 2D SS & 1,473 &8.2 &SU, C, SN, R & Front-view & \ding{51} & \ding{55} & GPS, IMU, CAN \\
		
		\bottomrule
		
	\end{tabular}
}

\label{tab:dataset_tab}
\end{table*}
 
	\section{Data Preparation}
	\label{data}
 
    In the context of autonomous driving, ensuring robustness in handling complex driving scenarios is paramount, given the high stakes involved in ensuring human safety. The autonomous driving system must effectively navigate diverse challenges, including traffic participants, weather conditions, lighting, and road conditions. However, it is impractical and inefficient to gather a dataset encompassing all possible scenarios, such as unexpected pedestrian-related traffic accidents.
    Furthermore, models trained on synthetic data may struggle to generalize effectively to real-world scenarios due to potential disparities in data distributions~\cite{}. Therefore, the crux of the issue lies in generating realistic and controllable data. Encouragingly, recent advancements, notably in diffusion models and NeRF, have produced images that blur the line between real and machine-generated, offering promising technical support for addressing data scarcity.
    
    This section delves into not only leveraging existing datasets but also exploring diverse approaches for collecting, synthesizing, or augmenting data in a cost-effective and efficient manner for autonomous driving. This encompasses techniques such as generative adversarial networks, diffusion models, neural radiance fields, and 3D Gaussian splatting. Tab.~\ref{tab:dataset_generation} provides an overview of these data generation methods.

	\subsection{Autonomous Driving Datasets}
	Data plays a pivotal role in training perception models for autonomous driving, particularly in establishing the foundational vision model for this domain. The evolution of automated driving technology over the past few decades has closely paralleled the improvement in dataset quality and richness.
	Tab.~\ref{tab:dataset_tab} provides a comprehensive overview of datasets pertaining to autonomous driving perception tasks from 2009 to 2024, encompassing details on sensor configurations, dataset diversity, and supported tasks. Notably, the dataset diversity includes various weather conditions, reflecting the real-world challenges encountered in autonomous driving scenarios.
	The table primarily focuses on widely used autonomous driving datasets, omitting annotation extensions such as \cite{tian2023occ3d}. For a more exhaustive survey of autonomous driving datasets, readers are directed to~\cite{li2023open,liu2024survey}.
	
    Autonomous driving datasets are typically gathered using vehicles equipped with a range of sensors, including cameras, LiDAR, radar, GPS, IMU, and CAN-bus. Each sensor type has its own strengths and weaknesses, necessitating their combined use to comprehensively capture environmental information. These sensors yield diverse data types such as RGB images, point clouds, millimeter wave radar data, GPS positioning information, and ego vehicle control attributes.
    
    The landscape of autonomous driving datasets is extensive, spanning academic and industrial domains. While most datasets feature RGB images, certain datasets like Caltech Pedestrian~\cite{dollar2009pedestrian}, Cityscapes~\cite{cordts2016cityscapes}, and IDD~\cite{varma2019idd} rely solely on cameras for data collection. Notably, these datasets exhibit varying camera configurations, capturing scenes from front-view perspectives~\cite{geiger2012we, alibeigi2023zenseact} to full $360^{\circ}$ surround views~\cite{caesar2020nuscenes, mao2021one}.
	
    LiDAR sensors, known for their ability to capture highly accurate point cloud data, are widely favored in autonomous driving datasets to ensure safety, resulting in the inclusion of 3D point cloud data in most datasets. In contrast, only a limited number of datasets provide radar data~\cite{caesar2020nuscenes}, despite radars' speed measurement capabilities and resilience to diverse weather conditions. This scarcity may stem from the processed nature of millimeter wave radar data, which poses challenges for deep learning due to its complex echo signal and non-trivial redundant information.
	
    In addition to real-world data, synthetic datasets such as Virtual KITTI~\cite{gaidon2016virtual}, Virtual KITTI 2~\cite{cabon2020virtual}, and the more recent UrbanSyn~\cite{gomez2023all} offer flexibility in simulating diverse weather conditions. However, they still grapple with the domain gap challenge~\cite{li2023intra}.

    \begin{table*}[t]
\centering
\caption{Overview of various data generation techniques in autonomous driving scenes. \textit{BBoxes} indicates bounding boxes. \textit{Cross-view} refers exclusively to those methods that are designed for generating multi-view images/videos. ``-" indicates that a field is inapplicable. You can find related datasets not included in Tab.~\ref{tab:dataset_tab} here: CARLA~\cite{dosovitskiy2017carla}, CARLA Town 01~\cite{pmlr-v78-dosovitskiy17a},  Gibson~\cite{xia2018gibson}, Udacity~\cite{udacity2016}, Brno Urban~\cite{ligocki2020brno} and PandaSet~\cite{xiao2021pandaset}. } 
\resizebox{\textwidth}{!}{
    \begin{tabular}{l|ccccc}
	\toprule
	{\textbf{Methods}} &  {\textbf{Categories}} & {\textbf{Dataset}} &\textbf{Input} & \textbf{Output} & \textbf{Cross-view}\\
	\hline
	
	Pix2PixHD~\cite{wang2018high} &   GAN &  Cityscapes & Images & Images & \ding{55}\\
	DeepRoad~\cite{zhang2018deeproad} &   GAN &  Udacity & Images & Images & \ding{55}\\
	
	LiDARGAN~\cite{sallab2019lidar} &   GAN &  KITTI/CARLA & LiDAR point clouds & LiDAR point clouds & - \\  
	
	DriveGAN~\cite{kim2021drivegan} &   GAN &  CARLA/Gibson & Videos, Actions & Videos & \ding{55}\\       
	
	\hline
	MCVD~\cite{voleti2022mcvd} &  Diffusion & Cityscapes & Images/Videos & Images/Videos & \ding{55}\\
	
	FDM~\cite{harvey2022flexible} &  Diffusion & CARLA Town 01 & Videos & Videos & \ding{55} \\
	
	GeoDiffusion~\cite{chen2023geodiffusion} &  Diffusion & nuScenes & Images/Videos, 2D BBoxes & Videos & \ding{51} \\
	
	DrivingDiffusion~\cite{li2023drivingdiffusion}  &  Diffusion & nuScenes  & 3D Layout  &  Videos &   \ding{51} \\
	
	DriveDreamer~\cite{wang2023drivedreamer}  &  Diffusion & nuScenes  & 3D BBoxes, Reference Image, HDMap, Actions   &  Videos &   \ding{51} \\
	
	MagicDrive~\cite{gao2023magicdrive}   &  Diffusion & nuScenes  & 3D BBoxes, Camera Pose, HDMap   &  Images  & \ding{51} \\
	
	
	\hline
	READ~\cite{li2023read} &  NeRF & KITTI/Brno Urban & LiDAR Point clouds (Obtain from Images) & Videos & \ding{51} \\
	
	MapNeRF~\cite{wu2023mapnerf} &  NeRF & Argoverse2 & Videos, Maps, Camera Poses & Videos & \ding{55} \\
	
	UniSim~\cite{yang2023unisim} &  NeRF & PandaSet & Videos, LiDAR point clouds, Camera Poses & Videos, LiDAR point clouds & \ding{51} \\
	
	MARS~\cite{wu2023mars} &   NeRF & KITTI  & Images, 3D BBoxes, Camera Poses   &  Images, Depth, Semantic &   \ding{51} \\
	
	LiDAR-NeRF~\cite{tao2023lidar} &   NeRF &  KITTI-360  &  LiDAR point clouds  &   LiDAR point clouds &   - \\
	
	NeRF-LiDAR~\cite{zhang2023nerf}  &   NeRF & nuScenes  & Images, LiDAR point clouds    &  LiDAR point clouds  &   -  \\
    \hline
    PVG~\cite{chen2023periodic} & 3DGS & KITTI/Waymo & Images & Images & \ding{51} \\
    DrivingGaussian~\cite{zhou2023drivinggaussian} & 3DGS & nuScenes/Waymo & Images, LiDAR point clouds (optional) & Images & \ding{51} \\
    Street Gaussians~\cite{yan2024street} & 3DGS & KITTI/Waymo & Images & Images & \ding{51} \\
    
	
	
	\bottomrule
	
\end{tabular}
}

\label{tab:dataset_generation}
\end{table*}
	\subsection{Generative Adversarial Network}
	
	Generative Adversarial Networks (GAN) have gained a lot of popularity since they were proposed in 2014~\cite{goodfellow2014generative}.
	The idea behind GAN is very simple and intuitive, consisting of two networks a generator and a discriminator. The generator's task is to generate samples, that are as similar to the real data samples as possible, while the discriminator tries to distinguish the real samples from the generated one. By optimizing these two networks like a two-player minimax game, the generator is able to synthesize photorealistic images.
	Thanks to the immense popularity of GAN, increasing applications in autonomous driving have been already identified~\cite{uricar2019yes,lehner20233d}.
    Here we mainly focus on GAN as an advanced data augmentation technique to synthesize realistic driving scenarios datasets.
	
	
	Pix2pix~\cite{isola2017image} and the subsequent work pix2pixHD~\cite{wang2018high} propose an image-to-image translation GAN, enabling the impressive synthesis of autonomous driving images by taking the semantic maps or edge images as conditional inputs. 
	%
	%
	Considering the mandatory requirements of paired images when training GAN in pix2pix, \cite{zhu2017unpaired} proposes an unpaired image-to-image translation algorithm named CycleGAN.
	Thus CycleGAN shows even more advanced examples of possible data augmentation, such as the CG to real or day to night or summer to winter translations.
	As a natural extension of the image-to-image translation, the video-to-video translation takes a further step towards generating temporal consistent video frames.
	For example, a spatio-temporal adversarial objective is used in~\cite{wang2018video} to synthesize 2k resolution videos of street scenes up to 30 seconds long.
	DeepRoad~\cite{zhang2018deeproad} leverages GAN to generate driving scenes with various weather conditions. They also use these generated images to test the consistency of DNN-based autonomous driving systems across different scenes. 
	DriveGAN~\cite{kim2021drivegan} introduces a novel GAN-based high-quality neural simulator for autonomous driving, that achieves the ability to control the weather as well as the locations of background objects.
	What's more, SufelGAN~\cite{yang2020surfelgan} proposes a simple yet effective data-driven approach for generating driving scenes, supporting the reconstruction of realistic camera images for novel positions and orientations of the self-driving vehicle and moving objects in the scene.
	
	Except for the image synthesis, GANs can be also used to generate realistic LiDAR point clouds.
	As a pioneering work, \cite{caccia2019deep} propose a GAN-based method that could generate high-quality LiDAR scans, and capture both local and global features of real lidar scans as well.
	LiDARGAN~\cite{sallab2019lidar} employ the CycleGAN to produce realistic LiDAR from simulated LiDAR (sim2real), as well as to generate high-resolution, realistic LiDAR from lower resolution one (real2real).
	Afterwards, \cite{lee2022gan} present a GAN-based LiDAR-to-LiDAR translation method, allowing the simulation of the point clouds data under various adverse weather conditions.

    While GAN-based methods can generate realistic images and LiDAR data, they are often constrained by relatively low resolution. Additionally, addressing the challenge of generating data with multi-sensor consistency presents a significant obstacle for GAN approaches.

     \begin{figure}[t]
		\begin{center}
			\includegraphics[width=0.9\linewidth]{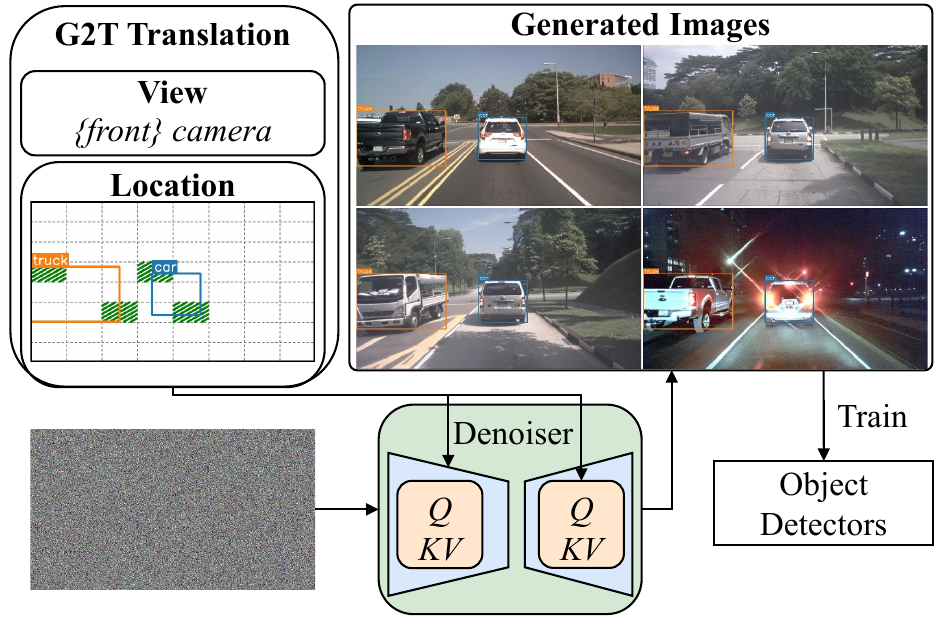}
		\end{center}
		
		\caption{\textbf{Illustration of diffusion-based data generation.}  The noise image, combined with the conditions of object bounding boxes' locations and geometry to text (G2T) description, is denoised into photo-realistic images. Images courtesy of \cite{chen2023geodiffusion}.}
		\label{fig:diffusion}
	\end{figure}
 
	\subsection{Diffusion Models}
	The diffusion models~\cite{2021Diffusion} have achieved remarkable success in image synthesis, surpassing the performance of GANs in certain respects. One key advantage lies in their use of high-scale classifier training on noisy images. Gradients extracted from this process guide the diffusion sampling towards specific class labels while maintaining a delicate balance between image fidelity and diversity.
    A typical diffusion-based data generation method pipeline is illustrated in Fig.~\ref{fig:diffusion}.
	As these earlier works well underway, subsequent work on conditional multi-frame or multi-view data generation exploded.
	%
	
	MCVD~\cite{voleti2022mcvd} utilized a probabilistic conditional score-based denoising diffusion model based on masking past and/or future frames in a sliding window blockwise autoregressive manner. 
	The mask-condition method was similarly used in Align your Latents~\cite{Blattmann_2023_CVPR} which additionally inserted temporal layers to force the model to align images in a temporally consistent manner in video generators. 
	FDM~\cite{harvey2022flexible} can be flexibly conditioned on an arbitrary number of frames through meta-learning. Besides that, a temporal attention mechanism including a position encoding network for generating long videos was explored.
	In addition to the works mentioned above, GeoDiffusion~\cite{chen2023geodiffusion} considered various geometric information including both multi-view and bounding box translation as prompt construction in the architecture.
	As for single frame data augmentation, DatasetDM~\cite{wu2023datasetdm} prepared a trainable decoder with few labeled images. Furthermore, it introduced a large language model that provides prompts to generate high-quality and infinitely numerous synthetic data for various downstream tasks. 
	
	As Bird's Eye View (BEV) and occupancy perception emerge as primary methodologies for autonomous driving perception, the generation of multi-view image sequences under the aforementioned conditions assumes growing significance.
    DriveDreamer~\cite{wang2023drivedreamer} employs a two-stage training pipeline. The initial stage focuses on understanding traffic structural information, while the subsequent video prediction training enhances predictive capabilities. This approach enables the controllable generation of driving scene videos that closely adhere to traffic constraints. Drive-WM~\cite{wang2023driving} introduces multiview and temporal modeling to jointly generate multiple views and frames. It predicts intermediate views conditioned on adjacent views through joint modeling factorization, thereby improving consistency between views. DrivingDiffusion~\cite{li2023drivingdiffusion} comprises a multi-view single-frame image generation model, a single-view temporal model, and post-processing modules that enhance cross-view and cross-frame consistency while extending the video length. Meanwhile, MagicDrive~\cite{gao2023magicdrive} serves as a single model, generating street view multiple camera images with multiple 3D geometry controls. Additionally, it incorporates a cross-view attention module to ensure consistency across multiple camera views.

      \begin{figure}[t]
		\begin{center}
			\includegraphics[width=0.9\linewidth]{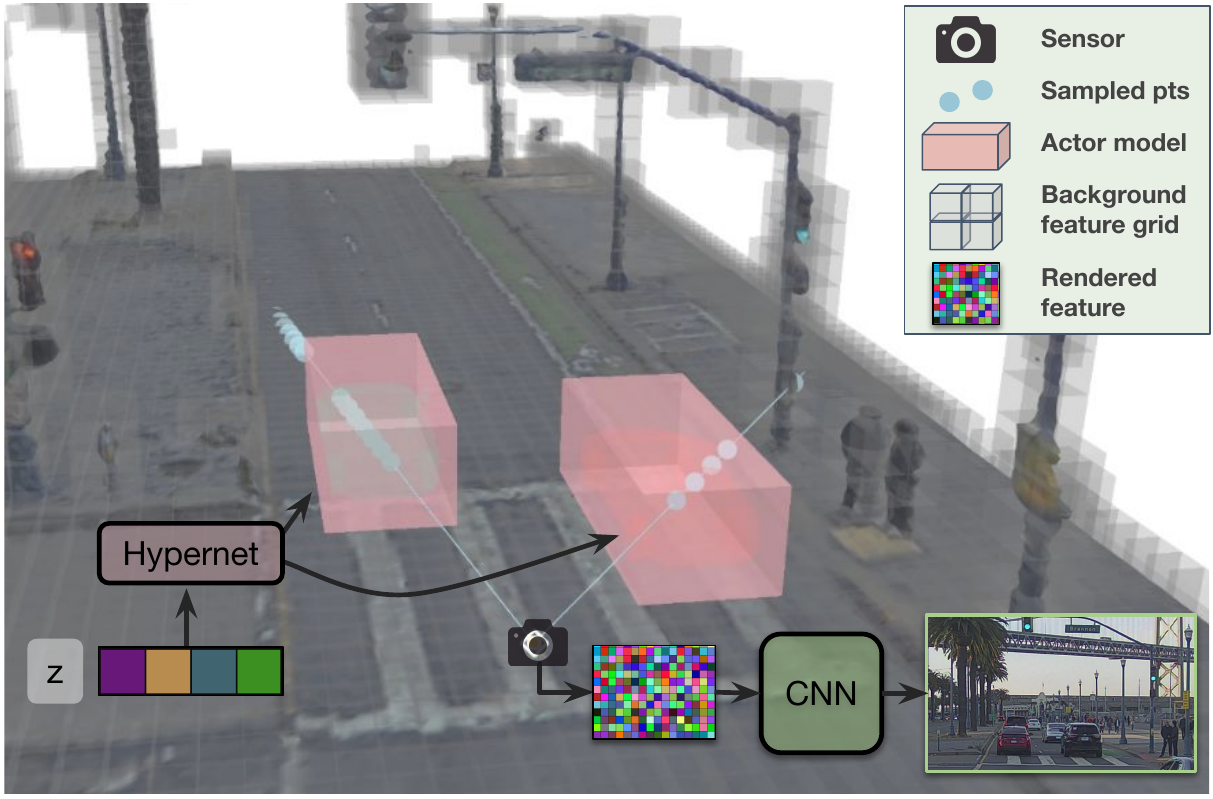}
		\end{center}
		
		\caption{\textbf{Illustration of NeRF-based data generation methods.} The 3D scene is modeled into a static background (grey) and a set of dynamic actors (red). The volume rendering is used to generate neural feature descriptors, followed by a convolutional network to decode feature patches into an image. Images courtesy of \cite{yang2023unisim}.}
		\label{fig:nerf}
	\end{figure}
 
    While diffusion-based data generation methods have been gaining increasing attention, they are challenging to train from scratch and heavily rely on pre-trained Stable Diffusion Models. Furthermore, they lack the capability to reconstruct 3D geometry.

	\subsection{Neural Radiance Field}
    While image-to-image translation methods using GANs or diffusion can synthesize photo-realistic street scenes, they struggle to generate novel views of scenes due to the absence of 3D constraints. Consequently, the Neural Radiance Field (NeRF)~\cite{mildenhall2021nerf} has emerged as a promising solution. First introduced by researchers at the University of California, Berkeley in 2020, NeRF diverges from traditional 3D reconstruction techniques, which represent scenes using explicit expressions like point clouds, grids, and voxels. Instead, NeRF samples each ray, capturing the 3D location of each sampling point and the 2D viewing direction of the ray. These 5D vector values are then input into a neural network to determine the color and volume density of each sampling point. NeRF constructs a field parameterized by a Multilayer Perceptron (MLP) neural network to continuously optimize parameters and reconstruct the scene, enabling high-quality novel view synthesis.
    Subsequently, a series of endeavors have been undertaken to adapt NeRF concepts to large-scale scenes, such as NeRF++~\cite{zhang2020nerfpp}, NeRF in the Wild~\cite{martin2021nerf}, Mip-NeRF~\cite{barron2021mip}, among others. Additionally, efforts have been made to train NeRF from a few input views, as demonstrated by works such as \cite{zhou2023single},  PixelNeRF~\cite{yu2021pixelnerf}, and Behind the Scenes~\cite{wimbauer2023behind}.
	
    Several approaches have been inspired by these works to leverage NeRF for simulating autonomous driving scenes. For instance, \cite{li2023read} introduces READ, a large-scale neural scene rendering method for autonomous driving. READ not only synthesizes realistic driving scenes but also facilitates the stitching and editing of driving scenes using neural descriptors. This method enables the synthesis of diverse driving scene data from different views, even for scenarios involving traffic emergencies.
    Furthermore, considering that collected images from driving scenes often exhibit similarity along the driving trajectory, which can lead to unsatisfactory outcomes, especially when the camera pose is positioned out-of-trajectory, MapNeRF~\cite{wu2023mapnerf} addresses this challenge by incorporating map priors such as ground and lane information in sampling computations. This incorporation guides the radiance field training, ultimately enhancing the semantic consistency of out-of-trajectory driving view synthesis.
    UniSim~\cite{yang2023unisim} leverages originally recorded sensor data from driving cars to create manipulable digital twins. The core concept involves constructing a compositional scene representation that accurately models the 3D world, encompassing dynamic actors and static scenes, as depicted in Fig.~\ref{fig:nerf}. Following training, UniSim demonstrates the capability to generate realistic, temporally consistent LiDAR and camera data from new viewpoints, facilitating the addition or removal of actors simultaneously.
    MARS (ModulAr and Realistic Simulator)~\cite{wu2023mars} is another neural sensor simulator that seeks to establish an open-sourced modular framework for photo-realistic autonomous driving simulation based on NeRFs.
    Furthermore, recent advancements such as S-NeRF~\cite{xie2023s}, RoMe~\cite{mei2023rome}, and StreetSurf~\cite{guo2023streetsurf} have also been developed for the reconstruction and simulation of large-scale driving scenes. These methods demonstrate significant potential for novel multi-view synthesis and scene editing.

    In addition to image synthesis, certain approaches are dedicated to simulating realistic LiDAR point clouds using NeRFs.
    LiDAR-NeRF~\cite{tao2023lidar} harnesses NeRF to enable the joint learning of geometry and the attributes of 3D points, eschewing the production of precise and regular LiDAR patterns through explicit 3D reconstruction or game engine simulation.
    Concurrently, NeRF-LiDAR~\cite{zhang2023nerf} makes use of real images and point cloud data gathered by self-driving cars to learn the 3D scene representation, point cloud generation, and label rendering. The resulting data is adept at enhancing the generation of substantial volumes of realistic LiDAR data for training models in autonomous driving.

    \begin{figure}[t]
		\begin{center}
			\includegraphics[width=\linewidth]{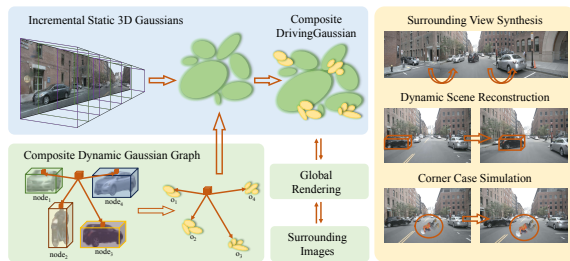}
		\end{center}
		
		\caption{\textbf{Illustration of 3DGS-based data generation methods.} The static background and dynamic objects are composited into explicit global 3D Gaussians. After training, it supports different data generation tasks, such as novel-view synthesis and corner case simulation. Images courtesy of \cite{zhou2023drivinggaussian}.}
		\label{fig:3dgs}
	\end{figure}

     NeRF has emerged as a compelling approach for multi-sensor consistent simulation, enabling the generation of photorealistic scenes that can be rendered from various viewpoints and with different light conditions.
    However, the current bottleneck for NeRF lies in the challenge of generating high-resolution data while meeting real-time processing requirements.
 
	\subsection{3D Gaussian Splatting}
    More recently, there has been a growing focus on 3D Gaussian Splatting (3DGS)-based methods~\cite{kerbl20233d}, which have garnered increased attention. Through 3DGS, scenes can be represented using 3D Gaussian primitives, enabling real-time rendering with minimal memory cost via rasterization-based rendering.
    
    Numerous approaches have been put forward for reconstructing driving scenes. PVG~\cite{chen2023periodic} introduces a Periodic Vibration Gaussian for large-scale dynamic driving scene reconstruction. By integrating periodic vibration, time-dependent opacity decay, and a scene flow-based temporal smoothing mechanism into the 3D Gaussian Splatting technique, PVG demonstrates superiority over NeRF-based methods not only in high-quality dynamic scene reconstruction and novel-view synthesis, but also in training and inference speed.
    Simultaneously, as depicted in Fig.\ref{fig:3dgs}, DrivingGaussian\cite{zhou2023drivinggaussian} hierarchically models complex driving scenes using sequential data from multiple sensors. The Incremental Static 3D Gaussians and Composite Dynamic Gaussian Graphs modules are employed to separately reconstruct the static background and multiple dynamic objects.
    In comparison to PVG, DrivingGaussian supports corner case simulation in real-world driving scenes by inserting arbitrary dynamic objects into the reconstructed Gaussian field while maintaining temporal coherence.
    Additionally, \cite{yan2024street} propose a novel Street Gaussians approach along with a tracked pose optimization strategy and a 4D spherical harmonics appearance model to handle the dynamics of moving vehicles. They also demonstrate that the proposed method allows for easy compositing of object vehicles and backgrounds, enabling scene editing and real-time rendering within half an hour of training.
	
    3D Gaussian Splatting enhances training speed and has the capacity to generate high-resolution images. Nonetheless, it is notable that 3DGS currently lacks the capability for comprehensive 3D scene representation, thereby suggesting a potential avenue for future research.
	
	\section{Self-supervised Training}
	\label{training}
    Upon acquiring extensive realistic data, effective pre-training paradigms are essential for extracting general information from massive datasets and constructing visual foundational models.
    
    Self-supervised learning, which involves training on large amounts of unlabeled data, has demonstrated promise in various domains, such as natural language processing and specific image processing applications. Furthermore, it has introduced new prospects for the development of VFMs for autonomous driving.
    As illustrated in Tab.~\ref{tab:ssl}, we conduct a comprehensive survey on the self training paradigms for constructing VFMs for autonomous driving, encompassing all endeavors in self-supervised or unsupervised manners. These methods are categorized into five primary types, including contrastive-based, reconstruction-based, distillation-based, rendering-based and world model-based.

    \begin{table*}[t]
\centering
\small
\caption{Overview of existing self-supervised learning (SSL) methods in autonomous driving scenes. } 
\resizebox{\textwidth}{!}{
\begin{tabular}{l|cccc}
\toprule
\textbf{Method}           &  \textbf{Model Input}      & \textbf{Type of SSL}    & \textbf{Pre-train Dataset} & \textbf{Vision Tasks} \\\hline
GCC-3D~\cite{liang2021exploring}           &  LiDAR        & Contrastive    & Waymo             & Detection       \\
ProposalContrast~\cite{yin2022proposalcontrast} &  LiDAR        & Contrastive    & Waymo             & Detection\\
SimIPU~\cite{li2022simipu}           &  LiDAR \& Camera & Contrastive    & KITTI             & Detection       \\
SegContrast~\cite{nunes2022segcontrast}      &  LiDAR        & Contrastive    & KITTI             & Segmentation       \\
BEVContrast~\cite{Sautier_3DV24}      &  LiDAR        & Contrastive    & KITTI/nuScenes    & Detection, Segmentation  \\
AD-PT~\cite{yuan2023ad-pt}            &  LiDAR        & Contrastive    & ONCE              & Detection       \\
\hline
BEV-MAE~\cite{lin2022bev}          &  LiDAR        & Reconstruction & Waymo             & Detection       \\
Voxel-MAE~\cite{min2022voxel}        &  LiDAR        & Reconstruction & Waymo             & Detection       \\
MAELi~\cite{krispel2024maeli}        &  LiDAR        & Reconstruction & Waymo             & Detection, Segmentation  \\
GD-MAE~\cite{yang2023gd}           &  LiDAR        & Reconstruction & Waymo             & Detection       \\
ALSO~\cite{boulch2023also}             &  LiDAR        & Reconstruction & KITTI             & Detection, Segmentation  \\
Occupancy-MAE~\cite{min2023occupancy}    &  LiDAR        & Reconstruction & ONCE              & Detection, Segmentation  \\
SPOT~\cite{yan2023spot}             &  LiDAR        & Reconstruction & Waymo             & Detection, Segmentation  \\
\hline
S2M2-SSD~\cite{zheng2022boosting}         &  LiDAR \& Camera & Distillation   & nuScenes          & Detection       \\
SLidR~\cite{sautier2022image}            &  LiDAR \& Camera        & Distillation   & nuScenes          & Detection, Segmentation  \\
ST-SLidR~\cite{mahmoud2023self}         &  LiDAR \& Camera        & Distillation   & nuScenes          & Segmentation       \\
SEAL~\cite{liu2023segment}             &  LiDAR \& Camera & Distillation   & nuScenes          & Segmentation       \\
\hline
PonderV2~\cite{zhu2023ponderv2}           &  LiDAR \& Camera       & Rendering      & nuScenes           & Detection, Segmentation  \\
PRED~\cite{yang2023pred}             &  LiDAR \& Camera & Rendering      & nuScenes \& ONCE  & Detection, Segmentation  \\
UniPAD~\cite{yang2023unipad}           &  LiDAR \& Camera & Rendering      & nuScenes          & Detection, Segmentation  \\\hline
UniWorld~\cite{min2023uniworld}           &  Camera & World Model      & nuScenes          & Detection, Occupancy  \\
ViDAR~\cite{yang2023vidar}           &  Camera & World Model      & nuScenes          & Det., Seg., Occ., Track., \textit{etc.}  \\
\bottomrule
\end{tabular}
}
\label{tab:ssl}
\end{table*}

	\begin{figure}[t]
		\begin{center}
			\includegraphics[width=\linewidth]{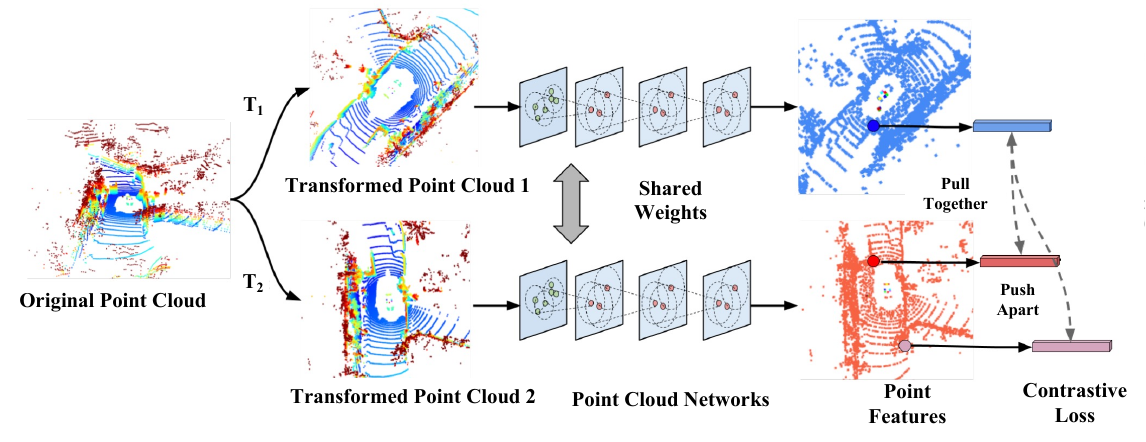}
		\end{center}
		
		\caption{\textbf{Illustration of scene-level contrastive learning-based methods.}  The original point cloud undergoes a transformation to produce two transformed point clouds. Images courtesy of \cite{shi2022self}. }
		\label{fig:contrastive}
		
	\end{figure}
	
	\subsection{Contrastive}
	Contrastive learning, exemplified by methods like MoCo~\cite{he2020momentum} and MoCov2~\cite{chen2020improved}, has emerged as a powerful tool for learning image representations by discriminating the similarities between augmented versions of the same image. This approach has yielded significant success in the 2D domain, and its potential has inspired researchers to explore its application to autonomous driving.
	
	\noindent\textbf{Scene-level Methods.} Pioneering works like PointContrast~\cite{xie2020pointcontrast} and DepthContrast~\cite{zhang2021self} leveraged view-based contrastive learning for 3D point clouds, aligning the features of two augmented point clouds, as depicted in Fig.~\ref{fig:contrastive}. However, these methods were primarily focused on indoor perception and lacked semantic information due to the limitations of their static partial view setting.
	To address these limitations, subsequent research, such as GCC-3D~\cite{liang2021exploring}, proposed a self-supervised learning framework that integrates geometry-aware contrast and clustering harmonization. By incorporating the prior knowledge that spatially close voxels tend to have similar local geometric structures, GCC-3D utilizes geometric distance to guide voxel-wise feature learning, alleviating the ``class collision" problem inherent in hard labeling strategies.
	Furthermore, \cite{li2022simipu} introduced SimIPU, a novel pre-training method designed specifically for outdoor multi-modal datasets. This method utilizes a multi-modal contrastive learning pipeline comprised of an intra-modal spatial perception component and an inter-modal feature interaction module, enabling the learning of spatial-aware visual representations.
	AD-PT~\cite{yuan2023ad-pt} treats point-cloud pre-training as a semi-supervised learning problem, effectively leveraging the few-shot labeled and massive unlabeled point-cloud data to generate unified backbone representations. This approach decouples the pre-training process and downstream fine-tuning task, making it directly applicable to many baseline models and benchmarks.
	
	\noindent\textbf{Region-level Methods.} Scene-level contrastive learning methods, while effective for capturing global context, can lead to the loss of crucial local details. 
	To address these limitations, region-based methods offer a compelling compromise. They strike a balance between global and local context, making them particularly suitable for 3D object detection and semantic segmentation tasks in diverse outdoor autonomous driving scenarios.
	To further enhance the performance of downstream semantic segmentation tasks, SegContrast pre-training~\cite{nunes2022segcontrast} extracts class-agnostic segments from the point cloud. These segments are then used to compute a segment-wise contrastive loss over augmented pairs, facilitating the learning of contextualized information crucial for accurate segmentation.
	For 3D object detection, \cite{yin2022proposalcontrast} proposed ProposalContrast, a novel two-stage proposal-level self-supervised learning framework. This framework leverages region proposals as the learning units, enabling the network to learn informative representations specifically tailored for object detection tasks.
	Instead of relying on computationally costly segmentation or proposal methods for point representation pooling, BEVContrast~\cite{Sautier_3DV24} takes a more efficient approach: projecting features onto the Bird's Eye View (BEV) plane and locally pooling them within 2D grid cells.


	\begin{figure}[t]
		\begin{center}
			\includegraphics[width=\linewidth]{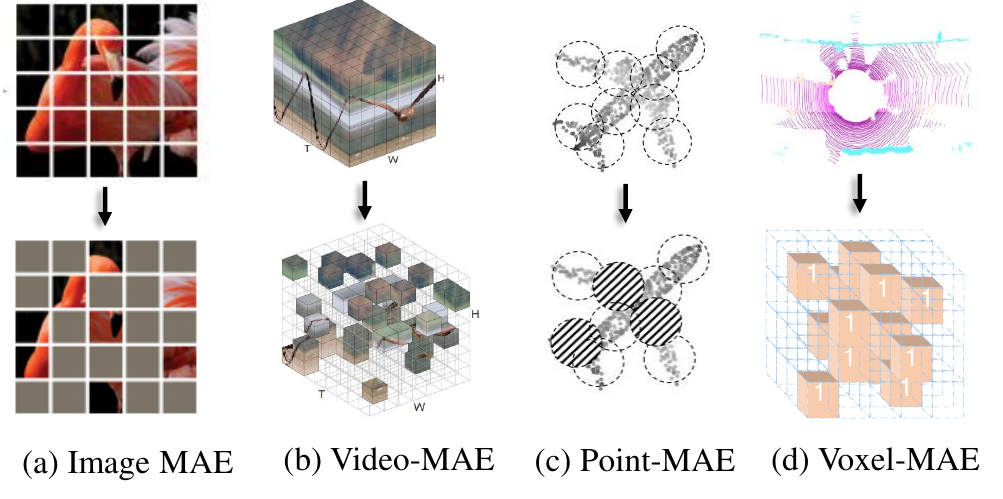}
		\end{center}
		
		\caption{\textbf{Illustration of reconstruction-based methods.} Masked Autoencoders for image (a), video (b), (c) synthetic point clouds, and (d) LiDAR point clouds. Images courtesy of \cite{min2022voxel}. }
		\label{fig:reconstruction}
		
	\end{figure}
	\subsection{Reconstruction}
	Reconstruction-based pre-training has emerged as a dominant force in the field of 3D perception for autonomous driving, encompassing both camera and point cloud-based approaches. This paradigm leverages self-supervised learning methods to pre-train perception models by reconstructing the input data from a masked or corrupted representation, as shown in Fig.~\ref{fig:reconstruction}.
	
	\noindent\textbf{Camera-based Reconstruction.} For camera-based perception models, methods leveraging Masked AutoEncoders (MAE) have demonstrated significant progress. Notably, MAE~\cite{he2021masked}, SimMIM~\cite{xie2022simmim}, MCMAE~\cite{gao2022mcmae}, MixMAE~\cite{liu2023mixmae} and SparK~\cite{tian2023designing} achieved impressive generalization capabilities by reconstructing masked image patches. These methods learn to reconstruct the masked regions, effectively encoding the underlying image patterns and relationships between various scene components. 
	GeoMIM~\cite{liu2023geomim}, the first to introduce camera-based MAE pretraining method into autonomous driving, which fully utilized the lidar bev features encoded by the pretrained bev model. Specifically, the lidar bev features are only used in training phase, not used in inference time, which seems like a distillation strategy for better geometry perception capability.

	\noindent\textbf{Point-cloud-based Reconstruction.} Point-cloud-based perception models also benefit significantly from reconstruction-based pre-training. PointMAE~\cite{pang2022masked} employs a set-to-set Chamfer Distance loss to restore masked points, ensuring accurate reconstruction while preserving the underlying 3D geometry. VoxelMAE~\cite{min2022voxel} takes a different approach, focusing on recovering the underlying geometry by differentiating occupied voxels. GeoMAE~\cite{tian2023geomae} introduce additional centroid, normal and curvature prediction tasks as prextext tasks to capture the geometric information of the point cloud. Alternatively, MaskPoint~\cite{liu2022masked} pre-trains the point cloud encoder through binary classification of occupied points.
	Building upon the foundation of reconstruction-based pre-training, \cite{krispel2024maeli} presented MAELi (Masked AutoEncoder for LiDAR). This innovative method leverages the inherent sparsity of LiDAR data by distinguishing between empty and non-empty voxels. Furthermore, it employs a novel masking strategy that specifically adapts to the unique spherical projection characteristics of LiDAR sensors, enabling effective reconstruction and representation learning.
	Seeking to address the challenge of occluded geometry, \cite{yang2023gd} introduced GD-MAE (Generative Decoder-based Masked AutoEncoder). This approach employs a generative decoder that automatically merges information from the surroundings in a hierarchical manner, effectively recovering occluded geometric knowledge and enhancing the overall representation of the scene.
	Additionally, BEV-MAE~\cite{lin2022bev} introduced a Bird's Eye View (BEV) strategy to guide the 3D encoder to learn feature representation from a BEV perspective, simplifying the pre-training process.
	With the advancement of 3D occupancy prediction, approaches like Occupancy-MAE~\cite{min2023occupancy}, ALSO~\cite{boulch2023also} and SPOT~\cite{yan2023spot} have emerged, focusing on pre-training point cloud backbones through 3D occupancy reconstruction.
	
	\noindent\textbf{Multi-modal Reconstruction.} Building perception models to learn from diverse, multimodal data remains an open challenge.
	PiMAE~\cite{chen2023pimae} introduces a self-supervised pre-training framework that promotes 3D and 2D  interaction in the mask tokens, which could boost performance of 2D and 3D detectors by a large margin. However, current works adopt a multi-stage pre-training system, where the complex pipeline may
	increase the uncertainty and instability of the pre-training. M3I~\cite{su2022allinone} propose a general multi-modal mutual information formula as a unified optimization target and demonstrate that all existing
	approaches are special cases of our framework. This approach achieves better performance than previous pre-training methods on various vision tasks, including classification, object detection and semantic segmentation. M3AE~\cite{geng2022multimodal} learns a unified encoder for both vision and language data via masked token prediction to learn generalizable representations that transfer well to downstream tasks.
	
	
		\begin{figure}[t]
		\begin{center}
			\includegraphics[width=\linewidth]{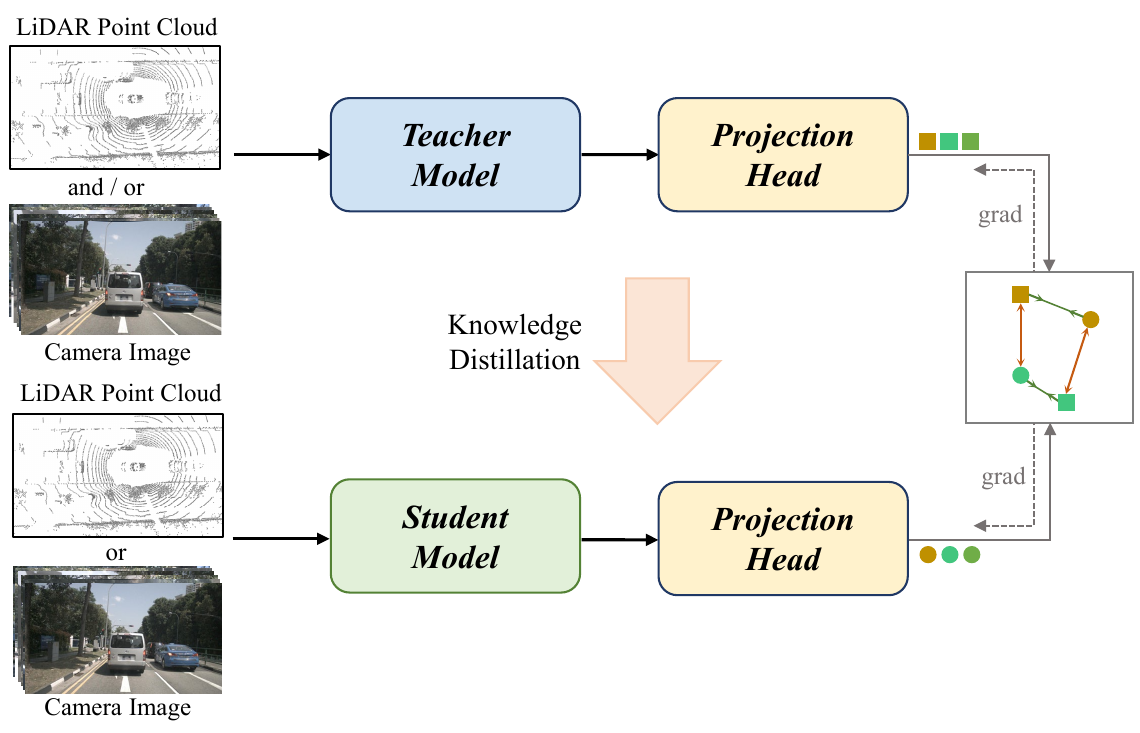}
		\end{center}
		
		\caption{\textbf{Illustration of distillation-based methods.} They mainly consist of the teacher model and the student model. The former is usually more powerful and can be pre-trained in large-scale datasets, it can accept either single-modality or multi-modality inputs. While the student model is usually more light-weight, and learned the knowledge from the teacher model via the contrastive loss or consistency constraints, thus enhancing the performances of the student model.}
		\label{fig:distillation}
		
	\end{figure}

	\subsection{Distillation}
	Distillation-based pre-training leverages knowledge gleaned from trained teacher backbones to enhance the performance of student counterparts~\cite{gou2021knowledge}, as shown in Fig.~\ref{fig:distillation}. For instance, some approaches capitalize on the strengths of image-based perception, where abundant data and established models exist, to improve the performance of LiDAR-based perception models, which often suffer from limited data and complex representations.
	
	SLidR~\cite{sautier2022image} and S2M2-SSD~\cite{zheng2022boosting} pioneered this field with their frameworks. they independently propose 2D-to-3D and fusion-to-3D representation distillation methods for cross-modal self-supervised learning on large-scale point clouds. These works demonstrated the promising potential of distillation for LiDAR pre-training, achieving significant performance improvements.
	Subsequent research has further refined and enhanced the SLidR pipeline. For example, the work of \cite{mahmoud2023self} introduced a semantically tolerant contrastive constraint and a class-balancing loss, leading to further performance gains.
	Most recently, SEAL~\cite{liu2023segment} builds upon the foundation of SLidR and proposes leveraging Vision Foundation Models SAM~\cite{kirillov2023segment} to establish the cross-modal contrastive objective. This approach leverages the powerful representations learned by VFMs to tackle the challenging task of cross-modal representation learning for LiDAR pre-training.
	
	These advancements highlight the effectiveness of distillation-based pre-training in improving the performance of LiDAR perception models. By leveraging knowledge from pre-trained image networks, researchers have achieved significant progress in tackling the downstream tasks associated with LiDAR data, paving the way for VFMs autonomous driving systems.
	

	\begin{figure}[t]
		\begin{center}
			\includegraphics[width=\linewidth]{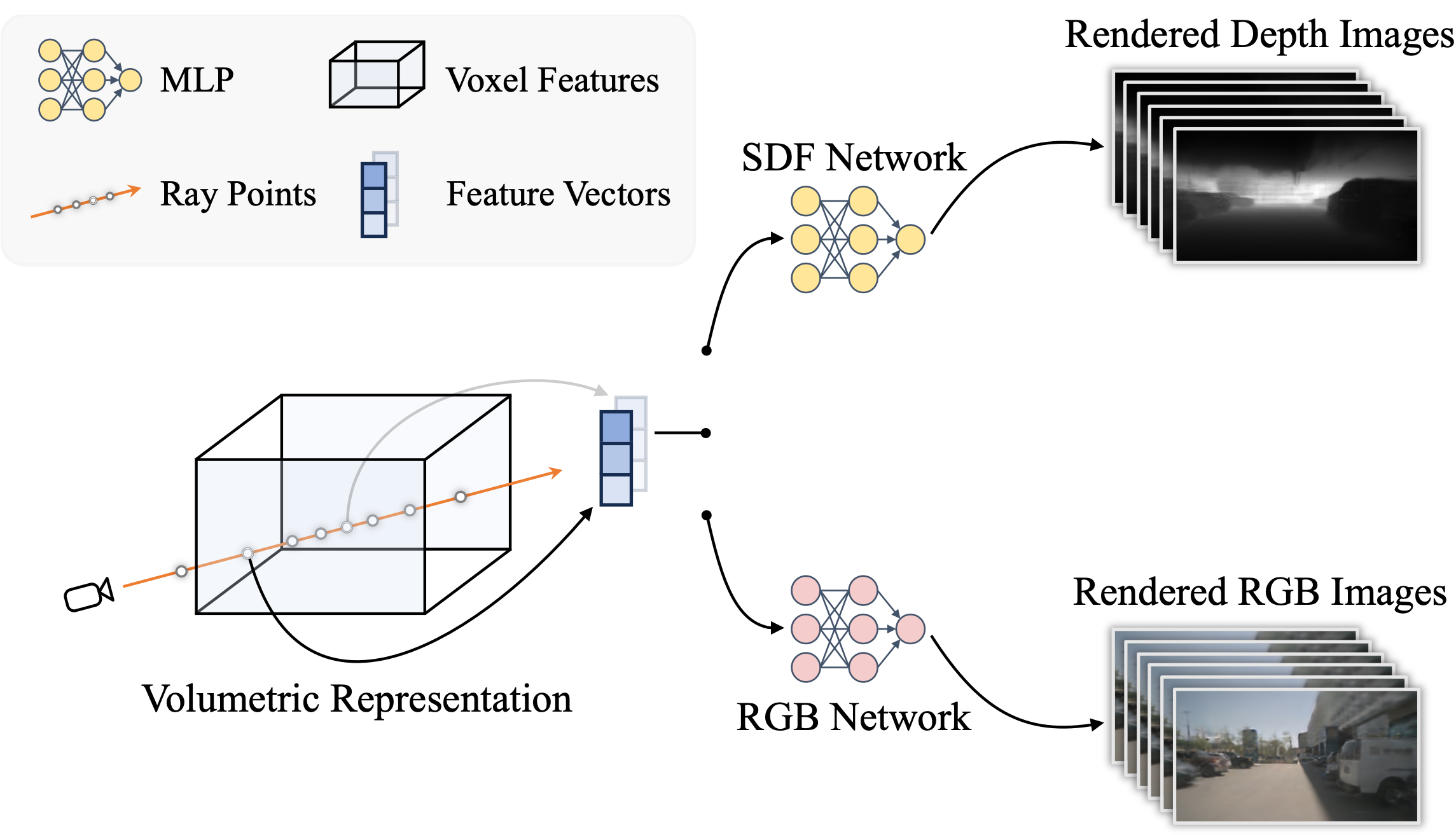}
		\end{center}
		
		\caption{\textbf{Illustration of rendering-based methods.}  The 3D feature volume is rendered to multi-view RGB images and depth maps via a differentiable neural rendering, which are compared with the original input multi-view RGB images and depths as the supervision. Images courtesy of \cite{min2022voxel}. }
		\label{fig:rendering}
		
	\end{figure}
	
	\subsection{Volume Rendering}
	Recent advancements have seen the emergence of rendering-based self-training approaches for autonomous driving perception. These methods operate by first mapping visual features, extracted from either point clouds or multi-view images, into a unified volumetric space. This allows for the incorporation of camera intrinsic and extrinsic parameters, facilitating the computation of corresponding rays for each pixel in each view image. Subsequently, MLPs are utilized to predict both the Signed Distance Function (SDF) and RGB values on sampled points along each ray, enabling differentiable volume rendering to reconstruct depth maps and images (see Fig.~\ref{fig:rendering} for the details).
	
	\noindent\textbf{Rendering for Pre-training.} Among these pioneering efforts, Ponder~\cite{huang2023ponder} stands as the first of its kind, laying the groundwork for this burgeoning field. However, its applicability is currently limited to indoor environments. In response, an advanced version~\cite{zhu2023ponderv2} has been proposed, extending the architecture through sparse voxel representation to encompass outdoor autonomous driving tasks, including object detection and semantic segmentation.
	PRED~\cite{yang2023pred} further leverages these techniques for LiDAR-based backbone pre-training. Notably, it incorporates a pre-trained semantic segmentation model to generate pseudo labels for rendered results, further enhancing pre-training effectiveness.
	Recently, UniPAD~\cite{yang2023unipad} offers a unified framework capable of accepting both multi-view images and point clouds as input. This framework utilizes a mask generator to partially mask the multi-modal inputs, enabling the network to focus on relevant information during the learning process.
	
	\noindent\textbf{Self-supervised Occupancy.} Beyond pre-training perception backbones, the rendering-based approach has also been explored as a complementary method for supervising occupancy prediction tasks. RenderOcc~\cite{pan2023renderocc} stands as the first to propose a method for occupancy prediction that relies solely on 2D supervision (depth and semantics maps). This method involves generating 3D rays across multiple frames, selecting rays through moving dynamic objects and employing class-balanced sampling, and ultimately rendering depth and semantic maps for supervision.
	SelfOcc~\cite{huang2023selfocc}, a concurrent work, adopts a similar training paradigm. However, it incorporates additional RGB supervision and leverages pseudo depth and semantic ground truths generated from pre-trained networks.
    More recently, OccNeRF~\cite{zhang2023occnerf} introduces a temporal photometric consistency loss to supervise the rendered depth, thereby eliminating the need for depth supervision.
	
	These rendering-based approaches have emerged as a powerful tool for VFMs development in autonomous driving. These methods offer advantages of utilizing vast 2D labels for 3D perception. 
	
	\begin{figure}[t]
		\begin{center}
			\includegraphics[width=\linewidth]{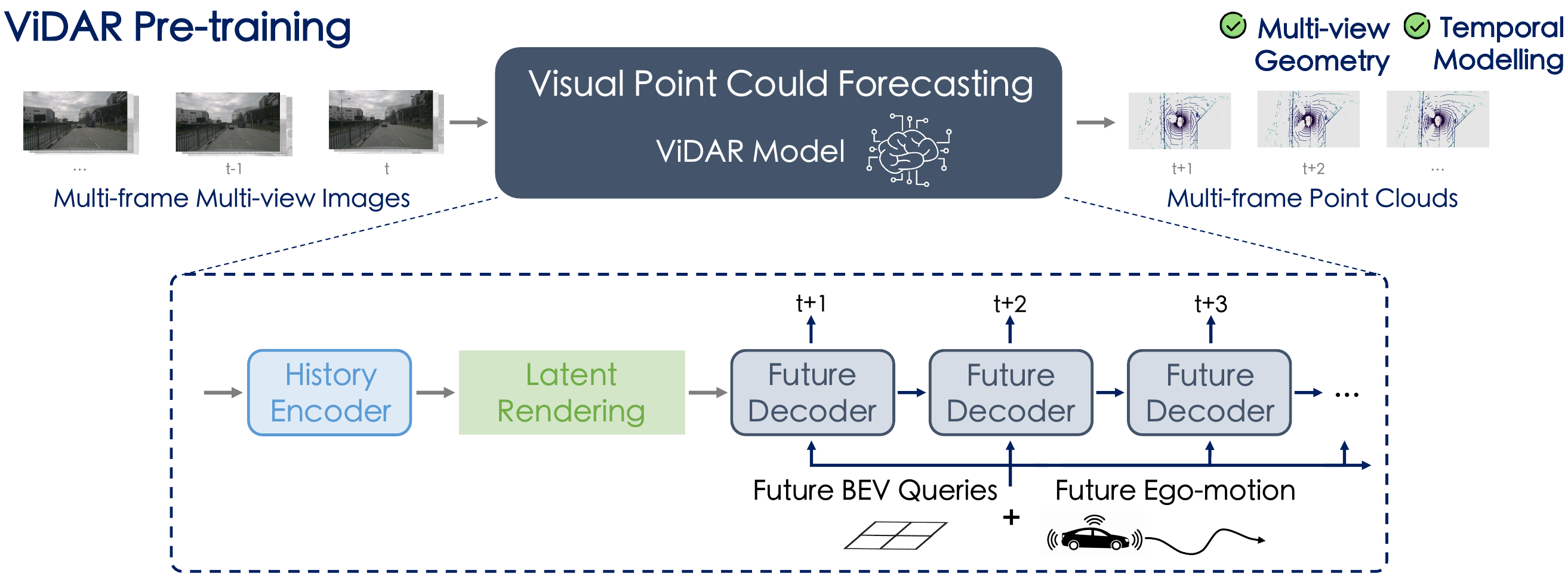}
		\end{center}
		
		\caption{\textbf{Illustration of world-model based methods.}  The prediction of future scans can serve as a pre-task for self-supervised learning. Images courtesy of \cite{yang2023vidar}.}
		\label{fig:world-model}
		
	\end{figure}
	
	\begin{table*}[t]
\footnotesize
\centering
\small
\caption{\textbf{Summary of techniques for world model.} We have involved different characteristics for each method, including input, output, encoder, decoder, and model architecture. \textit{I}, \textit{Act}, \textit{PC}, \textit{Traj}, and \textit{Occ} stand for images, action, point cloud, trajectory, and occupancy, respectively.} 
\resizebox{\textwidth}{!}{
    \begin{tabular}{l|cccccc}
    \toprule
        \textbf{Methods}  &{\textbf{Input}} &{\textbf{Encoder}} & \textbf{Architecture} & \textbf{Decoder} & \textbf{Output} \\ 
        \hline
        ADriver-I~\cite{jia2023adriveri} & I, Act (text format) & CLIP-ViT & Diffusion & Video Diffusion Decoder &  I, Act \\
        DriveDreamer~\cite{wang2023drivedreamer}  & I, Act, 3D box, Text & VAE & Diffusion & Video decoder, Action decoder &  I, Act  \\
        Driving-WM~\cite{wang2023driving} &  I, Act & VAE & Diffusion & VAE Decoder  & I, Traj\\
        OccWorld~\cite{zheng2023occworld} &  Occ, Ego Poses & VQVAE & GPT & VQVAE Decoder & Occ, Ego Poses \\
        Waabi-WM~\cite{zhang2023learning} &  PC, Act & VQVAE &  Discrete Diffusion & VQVAE Decoder & PC \\
        GAIA-1~\cite{hu2023gaia} &  I, Text, Act & VQVAE & GPT & Video Diffusion Decoder & I \\
        MUVO~\cite{bogdoll2023muvo} &  I, PC, Act & ResNet-18, SensorFusion & GRU & Task Specific Decoder & I, Act, Occ \\
        
        \bottomrule
        
    \end{tabular}
}

\label{tab:world_model}
\end{table*}
		
	\subsection{World Model}
	The world model is a long-standing concept in artificial intelligence, which is usually defined as predicting the future states conditioned on action and past observations~\cite{Ha_Schmidhuber}. 
	%
	%
	Owing to the abilities to allow the agent to reason about the surrounding world, predict future states, and make informed decisions without the need for consistent interaction with the real environment, they have achieved successful applications in the field of robotics, from simulation environments~\cite{schrittwieser2020mastering,hafner2023mastering} to real-world scenes~\cite{wu2023daydreamer,reed2022generalist}.
	
	For the autonomous driving task, the self-driving vehicles are traveling in dynamically changing scenes, which require the vehicles to possess knowledge about how the environment evolves.
	Therefore, world models for autonomous driving gained prominent attention in recent years~\cite{hu2023gaia,jia2023adriveri}, since they have the potential to pave the way to end-to-end autonomous driving~\cite{hu2023planning,jiang2022perceive}.
	From another perspective, the attempts at autonomous driving world models also introduced a promising training paradigm to forge the vision foundation models for autonomous driving.
	Specifically, world models are commonly trained in a self-supervised manner, which could be optimized in large-scale unlabeled data. Besides, by learning a generalizable representation of the world, agents can adapt to new tasks and challenges more easily.
	To this end, in this section, we mainly explore the works relating to autonomous driving world models, especially the approaches designed for representing and optimizing world models.
	The comparison of exiting world models are illustrated in Tab.~\ref{tab:world_model}.
	
	\noindent\textbf{Image-based World Model.} GAIA-1~\cite{hu2023gaia} proposes a generative world model for autonomous driving that takes video, text, and action as inputs, encoding them as a sequence of tokens.
	Then, the world model in GAIA-1 is designed as an auto-regressive transformer that predicts the next image tokens conditioned on all past encoded tokens.
	Finally, they utilize the video diffusion models~\cite{ho2204video} as the decoder to map the predicted image tokens back to the pixel space, greatly improving the temporal consistency of the output video.
	
	Although made a pioneering attempt at building a world model for autonomous driving, GAIA-1 is more like a generator for driving scenarios while ignoring the control signal prediction.
	Therefore, ADriver-I~\cite{jia2023adriveri} first presents the concept of infinite driving, by unifying the control signal prediction and the future scene generation.
	Given the historical vision-action pairs and current visual token as inputs, the ADriver-I can directly output the low-level control signals and the near future frames.
	To be specific, the multi-modal large language model (MLLM)~\cite{chiang2023vicuna,radford2021learning} reasons out the control signal of the current frame based on the inputs.
	After that, the predicted control signal, used as the prompt, together with the input tokens, are fed to the video diffusion model (VDM)~\cite{rombach2022high} to predict the future frames.
	This two-stage approach from DriveDreamer~\cite{wang2023drivedreamer} first learns to understand the underlying traffic structure, effectively building a mental map of the scene. In the second stage, it leverages this knowledge to predict future video frames, enabling controllable generation of driving scenarios that strictly adhere to traffic rules and regulations.
	Drive-WM~\cite{wang2023driving} tackles the challenge of multi-view and temporal coherence by jointly modeling multiple future views and frames. It then predicts intermediate views by conditioning them on adjacent ones, utilizing a factorization of the joint model. This technique significantly improves the visual consistency between generated views, resulting in more realistic and believable driving scene videos.

	\noindent\textbf{3D World Model.} OccWorld~\cite{zheng2023occworld} is a world model predict the movement of the ego car and the evolution of the surrounding scenes in the 3D Occupancy space. It first employ a VQVAE to refine high-level concepts and obtain discrete scene tokens in a self-supervised manner.  then tailor the generative pre-training transformers (GPT)  architecture and propose a spatial-temporal generative transformer to predict the subsequent scene tokens and ego tokens to forecast the future occupancy and ego trajectory.
	A world modeling approach is proposed in~\cite{zhang2023learning} by an AD startup Waabi (Waabi-WM in the table), which tokenizes sensor observations with VQVAE and then forecasts the future via discrete diffusion, as shown in Fig. 34. To decode and denoise tokens in parallel efficiently, Masked Generative Image Transformer (MaskGIT) is reformulated into the discrete diffusion framework with a few minor changes.
	Although the above work uses single-modal sensor data as input,  MUVO~\cite{bogdoll2023muvo} learns Geometric VOxel Representations  of the world by  utilizing raw camera and lidar data. And it can predict raw camera and lidar data as well as 3D occupancy representations multiple steps into the future, conditioned on actions. 
	
	\noindent\textbf{World Model for Pre-training.}
	UniWorld~\cite{min2023uniworld} takes a novel approach to pre-training by leverages massive unlabeled image-LiDAR pairs. Instead of relying on labels, it utilizes multi-view images as inputs, generating feature maps in a unified bird's-eye view (BEV) space. This BEV representation is then used to predict the occupancy of future frames through a world model head. The learned BEV features can then be transferred to and benefit other downstream tasks, even without explicit labels.
	ViDAR~\cite{yang2023vidar} focuses on predicting future point clouds based on past visual information, as depicted in Fig.~\ref{fig:world-model}. It first encodes historical frames into embedding vectors using an encoder network. These embeddings are then projected into 3D geometric space through a unique Latent Rendering operator, enabling the prediction of future point clouds. This approach shows significant performance improvements on over eight downstream tasks (\eg, 3D object detection, semantic segmentation, occupancy prediction, object tracking and future point cloud forecasting), highlighting the potential of world model pre-training for various vision tasks.

		\begin{table*}[t]
		\centering
		\small
		\caption{Overview of the adaptation of existing foundation models in autonomous driving scenarios. } 
		\resizebox{\textwidth}{!}{
			\begin{tabular}{l|lll}
				\toprule
				\textbf{Method} &  \textbf{Type} & \textbf{Task} & \textbf{Dataset} \\
				\hline
				CalibAnything~\cite{luo2023calib}&   VFM        & LiDAR-Camera Calibration   &KITTI    \\
				RobustSAM~\cite{shan2023robustness} &  VFM &2D Semantic Segmentation & BDD100k\\
				SPINO~\cite{kappeler2023fewshot} &  VFM & Panoptic Segmentation & Cityscapes/KITTI-360\\
				SEAL~\cite{liu2023segment} &VFM &Point Cloud Segmentation &SemanticKITTI/nuScenes/Waymo/\textit{etc.} \\
				RadOcc~\cite{zhang2023radocc}&  VFM & Occupancy Prediction & nuScenes\\
				\hline
    				GPT-Driver~\cite{mao2023gpt} &  LLM & Planning  &  nuScenes \\
				LanguageMPC~\cite{sha2023languagempc} &  LLM  &Decision-Making & IdSim~\cite{liu2021reinforcement} \\ 
				DrivelikeHuman~\cite{fu2024drive} & LLM &Planning & HighwayEnv~\cite{highway-env} \\
                PromptTrack~\cite{wu2023language} & LLM & Tracking & NuPrompt~\cite{wu2023language} \\
                HiLM-D~\cite{ding2023hilm} & LLM & Scene Recognition, Decision-Making & DRAMA~\cite{malla2023drama} \\
                DriveGPT4~\cite{xu2023drivegpt4} & LLM & Scene Recognition, Decision-Making & BDD-X \\
                LiDAR-LLM~\cite{yang2023lidar} & LLM & Scene Understanding, Planning, Grounding & nu-Caption~\cite{yang2023lidar} \\\hline
				CLIP2Scene~\cite{chen2023clip2scene} &  CLIP & Point Cloud Semantic Segmentation & SemanticKITTI/nuScenes/ScanNet \\
				OVO~\cite{tan2023ovo} &  CLIP & 3D Occupancy Prediction & NYUv2/SemanticKITTI \\
				POP-3D~\cite{vobecky2023pop} &  CLIP & 3D Occupancy Prediction & nuScenes \\
				Dolphins~\cite{ma2023dolphins} &  VLM &Scene Recognition, Planning &BDD-X~\cite{kim2018textual}\\
				On the Road with GPT-4V~\cite{wen2023road} &  VLM & Scene Recognition, Decision-Making & nuScenes/CARLA/DAIR-V2X/BDD-X \\
                DriveLM~\cite{sima2023drivelm} & VLM & Scene Understanding, Planning, Decision-Making & DriveLM-Data~\cite{sima2023drivelm} \\
                Reason2Drive~\cite{nie2023reason2drive} & VLM & Scene Understanding, Planning, Decision-Making & nuScenes/Waymo/ONCE \\
                
				\bottomrule
			\end{tabular}
		}
    \label{tab:foundation_model}
	\end{table*}
	
	\section{Adaptation}
	\label{adapt}
	%
    While the current lack of a tailored vision foundation model for self-driving presents a challenge, it is possible to analyze the application of existing foundation models, such as Vision Foundation Model, Multimodal Foundation Models, and Large Language Models from other fields, to enhance our understanding. 
    Tab.~\ref{tab:foundation_model} provides a clear summary of some prominent models. By examining the limitations of existing solutions, we have extracted key insights and proposed a dedicated Visual Foundation Model customized for autonomous driving.

	\subsection{Vision Foundation Model}

	%
	%
	The community has enthusiastically embraced vision foundation model, (\eg, SAM and DINO), leading to a wave of research exploring its potential. Extensions delve into diverse applications like image inpainting~\cite{yu2023inpaint}, image captioning~\cite{wang2023caption}, video object tracking~\cite{yang2023track}, and medical image analysis~\cite{ma2023segment,huang2023segment}.
	Expanding beyond 2D, subsequent research like~\cite{shen2023anything,zhang2023sam3d,chen2023ma,yang2023sam3d,zhou2023openannotate3d} leverage these VFMs prowess in 3D analysis tasks.
	
	Several recent works have explored the potential of adapting existing VFMs for various autonomous driving challenges, seeking to leverage existing vision foundation models in this new domain.
	Specifically, Calib-Anything~\cite{luo2023calib} utilizes SAM to design a LiDAR-camera calibration method that requires zero extra training and adapts to common scenes. 
	\cite{shan2023robustness} investigates SAM's segmentation robustness under adverse weather conditions. 
	SPINO~\cite{kappeler2023fewshot} leverages task-agnostic image features extracted from DINOv2~\cite{oquab2023dinov2} to enable few-shot panoptic segmentation. It showcases its generalizability by successfully applying it to different AD datasets.
	
	Moreover, VFM are also widely used during training to enhance model performance, extending their impact beyond direct segmentation tasks.
	For instance, SEAL~\cite{liu2023segment} pioneers the use of SAM-like models for self-supervised representation learning on large-scale 3D point clouds. 
	\cite{peng2023sam} utilizes instance masks generated by these models to improve the performance of unsupervised domain adaptation in 3D semantic segmentation. 
	RadOcc~\cite{zhang2023radocc} utilizes SAM to provide shape priors and performs segment-guided affinity distillation, leveraging the cross-modal knowledge transfer to enhance feature representations of 3D occupancy.
	
    Although these VFMs demonstrate proficiency in image-based perception tasks, they exhibit limitations in capturing 3D information. Furthermore, due to their customized architecture, integrating newly incoming modalities (\eg, LiDAR point cloud) as input poses a significant challenge.
 
    \begin{figure}[t]
		\begin{center}
			\includegraphics[width=0.9\linewidth]{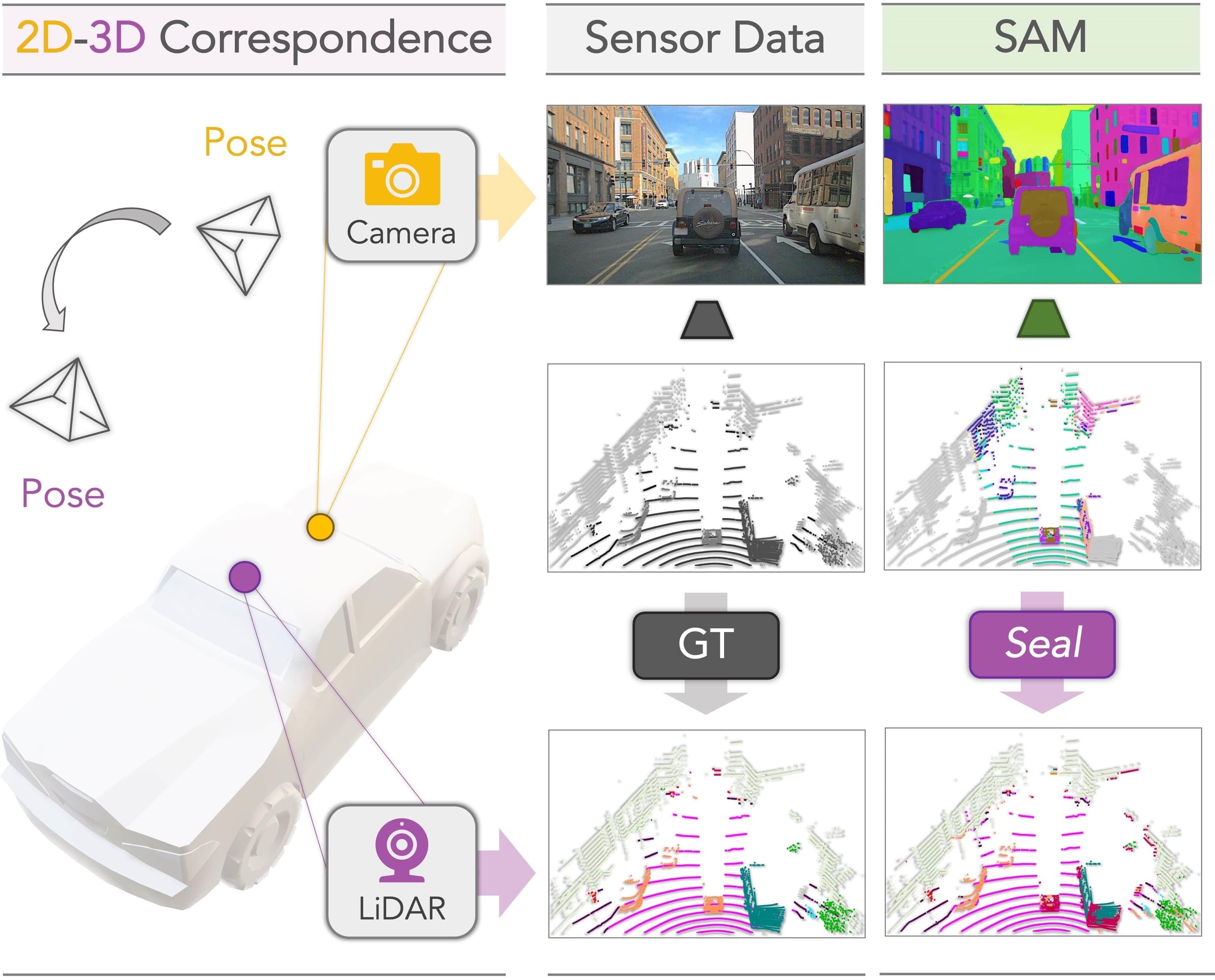}
		\end{center}
		
		\caption{\textbf{Illustration of a method that adapts SAM to segment any point cloud.} Images courtesy of \cite{liu2023segment}.}
		\label{fig:sam}
		
	\end{figure}
	
		
		

	\subsection{Large Language Models}

	The capabilities of Large Language Models (LLM) for generalization and interpretability are attracting significant attention in the autonomous driving community. Their proficiency in few-shot learning enables them to effectively handle out-of-distribution scenarios, such as encountering rare objects. Additionally, the inherent reasoning ability of LLMs renders them highly suitable for tasks that require logical processing and decision-making.
	
\begin{figure}[t]
		\begin{center}
			\includegraphics[width=\linewidth]{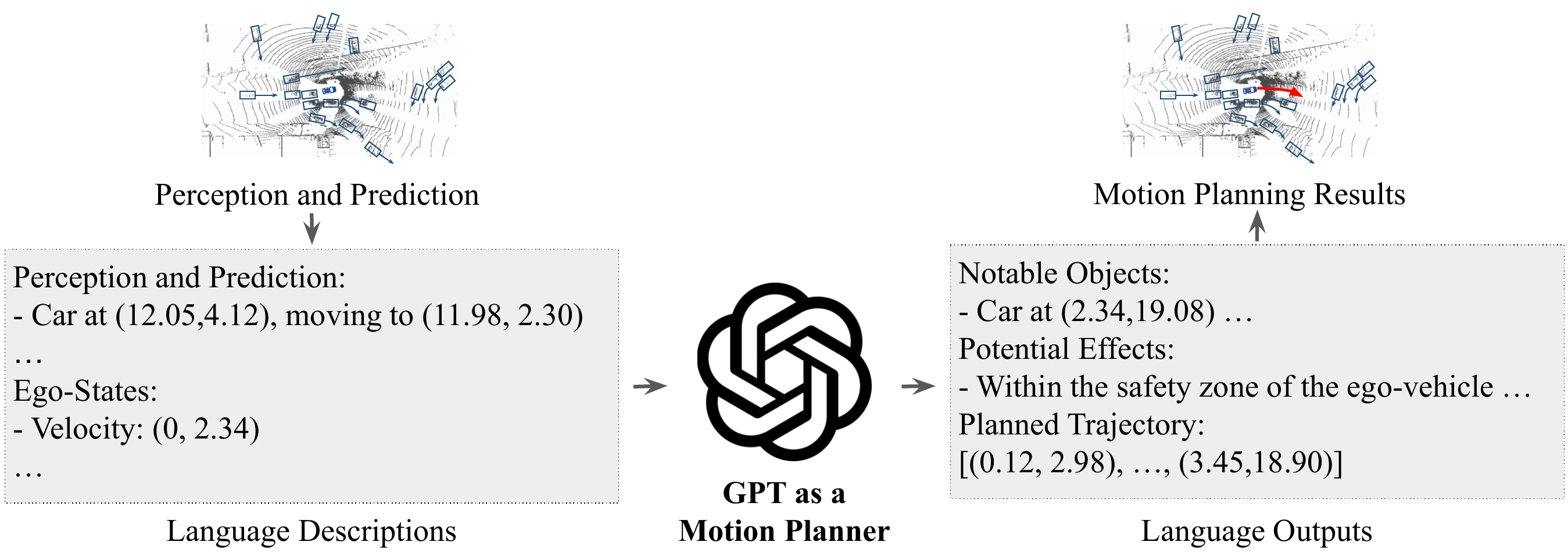}
		\end{center}
		
		\caption{\textbf{Illustration of a method that adapts the large language model as a motion planner for autonomous driving}. Images courtesy of \cite{mao2023gpt}.}
		\label{fig:adap_llm}
		
	\end{figure}

	\noindent\textbf{Planning.} LLM has been applied to generate control signals and explain the driving policy. A typical pipeline is shown in Fig.~\ref{fig:adap_llm}, the GPT-Driver~\cite{mao2023gpt} applies GPT3.5 as a motion planner for improved trajectory generation. Drive Like a Human~\cite{fu2024drive} uses GPT3.5 to explore LLMs in driving scenarios, focusing on human-like reasoning, interpretation, and problem-solving in long-tail cases.  LanguageMPC~\cite{sha2023languagempc} integrates LLMs with Model Predictive Control for enhanced decision-making. DiLu~\cite{wen2023dilu} utilizes GPT3.5 and 4 for a knowledge-driven system focusing on reasoning, reflection, and memory.
	
	However, current LLMs is deemed inadequate for comprehensive driving tasks~\cite{Chen2023-je}. A primary limitation is their inability to fully understand 3D space, a critical requirement for tasks like parking. The exploration of how to expand current LLMs into foundation model for driving remains an open and exciting research direction.

	\noindent\textbf{Perception.} This survey also emphasizes the potential of LLMs to serve as the fundamental building blocks for vision-based autonomous driving systems. LLMs excel in adapting to diverse vision tasks, particularly in data-scarce environments, where their few-shot learning capabilities enable rapid and accurate model adaptation and reasoning.
	PromptTrack~\cite{wu2023language} proposes a novel approach that fuses cross-modal features through a prompt reasoning branch to predict the 3D locations and motions of objects. It leverages semantic cues embedded in language prompts to effectively combine LLMs with existing 3D detection and tracking algorithms.
	HiLM-D~\cite{ding2023hilm} introduces a high-resolution multimodal LLM architecture specifically designed for the challenging task of Risk Object Localization and Intention and Suggestion Prediction. By incorporating fine-grained visual information into the LLM framework, HiLM-D improves the model's ability to localize potential hazards and predict the intentions and potential actions of other agents within the driving scene.
     DriveGPT4~\cite{xu2023drivegpt4} focuses on building vision question-answering capabilities tailored to the specific needs of autonomous driving. The model is trained on diverse scene-related questions, encompassing aspects like vehicle states, navigational guidance, and traffic situation comprehension, enabling it to provide interpretable and context-aware responses to queries encountered during autonomous operation.
    More recently, LiDAR-LLM~\cite{yang2023lidar} shows the potential of LLMs for 3D LiDAR understanding. The key insight of LiDAR-LLM is the reformulation of 3D outdoor scene cognition as a language modeling problem, enabling 3D QA, and zero-shot planning tasks.
    In these applications, the benefits of LLMs do not directly impact the downstream task. Furthermore, their performance has not reached the state-of-the-art.
    
	\subsection{Multimodal Foundation Models}
    Given the utilization of multi-modal sensors in autonomous driving, Multimodal Foundation Models present a viable solution for future research.
    
	\noindent\textbf{CLIP.}  Contrastive Language-Image Pre-training (CLIP) \cite{radford2021learning} is a foundational building block for multi-modal learning in computer vision. This large-scale model, based on the Transformer architecture, comprises a visual and a textual encoder that independently process input images and captions, respectively. Pre-trained on 400 million (image, text) pairs from the web, CLIP achieves state-of-the-art image representation performance. The alignment score between an image and text is computed as the dot product of the encoders' outputs. Building upon CLIP's success, the research community has seen a surge of extensions exploring its capabilities and applying it to diverse downstream tasks, including regression~\cite{yao2022detclip,yao2023detclipv2}, retrieval~\cite{luo2021clip4clip,fang2021clip2video}, generation~\cite{mokady2021clipcap,hong2022avatarclip}, segmentation~\cite{yu2023convolutions,tan2023ovo}, and others~\cite{Zhou_2022,zhou2022conditional,gao2023clip}. In this section, we focus on the promising applications of CLIP in autonomous driving perception.
	
	Open-vocabulary semantic segmentation poses a crucial challenge for autonomous driving perception since it allows the model to recognize the newly incoming objects. Several recent works leverage CLIP's pre-trained text embeddings to tackle this problem. LSeg~\cite{li2022languagedriven} utilizes these embeddings to learn pixel-level features for effective segmentation. MaskCLIP~\cite{dong2023maskclip} bypasses the self-attention pooling layer, generating pixel-level feature maps and employing text embeddings to predict the final segmentation mask. FC-CLIP~\cite{yu2023convolutions} leverages a frozen convolutional CLIP to predict class-agnostic masks and utilizes mask-pooled features for classification. ODISE~\cite{xu2023openvocabulary} takes a generative approach, employing a text-to-image diffusion model to propose mask candidates and perform classification. To further enhance open-vocabulary performance, ODISE classifies masks using features cropped from pre-trained CLIP.

    \begin{figure}[t]
		\begin{center}
			\includegraphics[width=\linewidth]{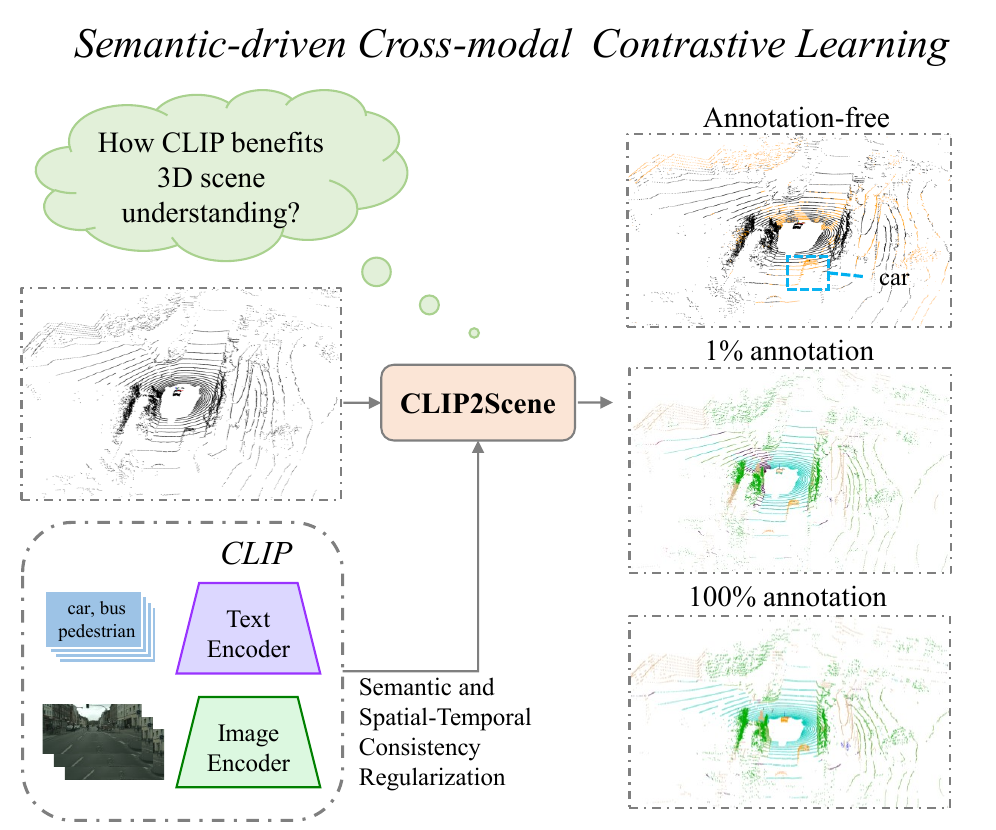}
		\end{center}
		
		\caption{\textbf{Illustration of a method that adapts a multi-modal foundation model \ie CLIP for 3D scene understanding}. Images courtesy of \cite{chen2023clip2scene}.}
		\label{fig:adap_clip}
		
	\end{figure}
 
	CLIP's potential extends beyond 2D perception. As shown in Fig.~\ref{fig:adap_clip}, CLIP2Scene~\cite{chen2023clip2scene} introduces a 2D-3D calibration matrix, enabling the application of MaskCLIP~\cite{dong2023maskclip} to 3D scene understanding tasks. OVO~\cite{tan2023ovo} offers a novel approach for semantic occupancy prediction of arbitrary classes without requiring 3D annotations during training. This strategy leverages knowledge distillation from a pre-trained 2D open-vocabulary segmentation model to the 3D occupancy network and employs pixel-voxel filtering for high-quality training data generation. Lastly, POP-3D~\cite{vobecky2023pop} proposes a framework for predicting open-vocabulary 3D semantic voxel occupancy maps from 2D images. This challenging problem, encompassing 2D-3D ambiguity and open-vocabulary nature, opens new avenues for 3D grounding, segmentation, and retrieval with free-form language queries.
	
	\noindent\textbf{VLMs.} While CLIP has emerged as a prominent force in multi-modal learning for autonomous driving, its reign is not unchallenged. Other emerging VLMs are demonstrating noteworthy perceptual capabilities and hold significant potential for end-to-end driving applications. Notably, Dolphins~\cite{ma2023dolphins} leverages OpenFlamingo to amplify the reasoning and interactivity of its autonomous driving system. This approach showcases the ability of VLMs to go beyond pure image/text alignment and incorporate higher-level cognitive functions into the decision-making process. Similarly, On the Road with GPT-4V~\cite{wen2023road} presents a compelling study, rigorously testing a state-of-the-art VLM on critical tasks for autonomous driving such as rare object detection, causal reasoning, and decision-making under uncertain scenarios. 
	Moreover, the VLMs are applied in diverse question-answering tasks of AD. 
	For instance, DriveLM~\cite{sima2023drivelm} recognized that human drivers reason about decisions in multiple steps rather than the single-round VQA, so it introduces Graph Visual Question Answering (GVQA) to mimic the human reasoning process. To validate the effectiveness of this approach, they create the DriveLM-nuScenes and DriveLM-CARLA datasets and provide a challenging benchmark for this task.
	Meanwhile, Reason2Drive~\cite{nie2023reason2drive} collects chained question-answer pairs about a sequential combination of perception, prediction, and reasoning steps, from open-source driving datasets, such as nuScenes, Waymo, and ONCE. Built upon the collected datasets, they introduce an interpretable and chain-based reasoning autonomous system.
	These works highlight the evolving landscape of VLMs in autonomous driving, suggesting that CLIP may not be the sole protagonist in this domain. Continued exploration and comparison of diverse VLM architectures and learning paradigms will be crucial in unlocking the full potential of multi-modal perception for robust and reliable autonomous driving solutions.

	\section{Present and Future}
	Driven by the methodology outlined above, we conclude present trends and propose several vital research directions with the potential to significantly advance the field of forging vision foundation models for autonomous driving.
	
	\subsection{Data Preparation}
	\subsubsection{Data Collection}
	The evolution of autonomous driving datasets can be demarcated into two distinct generations. The first, exemplified by KITTI~\cite{geiger2012we}, is marked by limited sensor modalities, relatively small data volumes, and a focus on perception-level tasks. The second generation, spearheaded by datasets like nuScenes~\cite{caesar2020nuscenes} and Waymo~\cite{sun2020scalability}, exhibits a significant leap in sensor complexity, data scale and diversity, and task scope, encompassing not only perception but also prediction and control~\cite{li2023open}.
	Looking ahead, the future of autonomous driving datasets may lie in leveraging the vast potential of unlabeled data, derived from both online sources and the ongoing operational deployments of self-driving vehicles.
	
	\subsubsection{Data Generation}
	The evolution of data generation algorithms for autonomous driving exhibits a clear trajectory toward multifaceted data synthesis. Early algorithms primarily focused on generating singular data modalities, such as LiDAR scans or camera images. However, the latest generation demonstrates advanced capabilities:
	\begin{itemize}
		 \setlength{\itemsep}{0pt}
		 \setlength{\parsep}{0pt}
		 \setlength{\parskip}{0pt}
		\item[-] \textbf{Multi-modal Consistency}: These algorithms can generate data across multiple modalities, ensuring coherence and inter-sensor validity.
		\item[-] \textbf{Enhanced Scenario Simulation:} The newest algorithms can manipulate specific elements within the virtual environment. This allows for the insertion of challenging, ``corner case" scenarios, such as pedestrians appearing suddenly or vehicles malfunctioning. 
		\item[-] \textbf{Diverse Driving Conditions:} Advanced algorithms can now incorporate a wider range of environmental factors into the generated data. This includes variations in weather (fog, rain, snow), lighting (night driving, sun glare), and even seasonal changes. 

	\end{itemize}
			
	Looking ahead, research in data generation may pivot towards leveraging the emerging field of Artificial Intelligence Generative Computing (AIGC). 
	Instead of solely generating brand new data, AIGC algorithms could modify and augment existing datasets. This would allow for efficient data expansion and customization, tailoring training sets to specific geographical regions or driving scenarios.

	\subsection{Self-supervised Training}
	Self-supervised learning has become a vital technique for enhancing autonomous driving models without the need for extensive labeled data. This approach is rapidly evolving, exhibiting several key trends:
		\begin{itemize}
		\setlength{\itemsep}{0pt}
		\setlength{\parsep}{0pt}
		\setlength{\parskip}{0pt}
		\item[-] \textbf{From Single-Modal to Multi-Modal:} Early self-supervised methods primarily leveraged single sensor modalities, like camera images or LiDAR scans. However, the latest generation embraces multi-modality.
		\item[-] \textbf{Multi-View and Temporal Consistency:} The newest approaches integrate multi-view and temporal consistency. 
		\item[-] \textbf{Learning 3D from Images:} Another exciting trend is the ability to utilize image information to infer 3D shape priors. This allows training 3D models directly from unlabeled images, eliminating the need for expensive 3D data.
		\end{itemize}
	
	Future research can focus on further strengthening the interactions between different modalities, enabling seamless information exchange and joint reasoning across sensors. 
	Moreover, exploiting knowledge distillation from powerful foundation models trained on massive datasets, like large language models (LLMs), is another intriguing direction. This could accelerate the learning process for specialized autonomous driving tasks and potentially unlock new capabilities.
	
	\subsection{Adaptation}
	Vision foundation models are still nascent in the autonomous driving domain, prompting researchers to leverage existing FMs from other areas for adaptation. This adaptation takes three main forms:
	\begin{itemize}
		\setlength{\itemsep}{0pt}
		\setlength{\parsep}{0pt}
		\setlength{\parskip}{0pt}
		\item[-] \textbf{Prior Extraction:}  Many approaches utilize FMs like SAM or DINO to extract informative image patches, which are then used as priors for downstream tasks like object detection or segmentation. 
		\item[-] \textbf{Hybrid Architectures:} Some world models integrate pre-trained FMs (\eg, GPT) as components to predict future sequences of tokens representing the driving environment. This leverages the reasoning capabilities of the LLMs to enhance the model's prediction.
		\item[-] \textbf{Knowledge Distillation and Assisted Training:} Techniques like knowledge distillation transfer the compressed knowledge of an FM to a smaller AD model, improving its performance without requiring excessive resources. Further research explores using FMs to guide the training process of new AD models.
	\end{itemize}
	
	Looking ahead, several key challenges and opportunities lie ahead in the development of VFMs for AD:
	\begin{itemize}
	\setlength{\itemsep}{0pt}
	\setlength{\parsep}{0pt}
	\setlength{\parskip}{0pt}
	\item[-] \textbf{Multi-Task Fine-Tuning:}  A central issue is how to effectively fine-tune a VFM on multiple downstream tasks (\eg, object detection, trajectory prediction) simultaneously. This requires algorithms that can coordinate task-specific adaptations while maintaining shared visual representations.
	\item[-] \textbf{Synergistic Task Interactions:} Research needs to explore how different tasks involving diverse sensory modalities can cooperate and mutually enhance each other within the VFM framework. This could lead to models with a more holistic understanding of the driving environment.
	\item[-] \textbf{Real-Time Deployment:} Due to the critical real-time requirements of AD, another crucial trend is developing efficient VFMs that can run effectively on onboard hardware within the vehicle. This involves exploring model compression techniques, lightweight architectures, and specialized hardware acceleration.
	\end{itemize}
	By addressing these challenges and capitalizing on emerging trends, VFMs have the potential to revolutionize the field of AD. They offer the promise of robust, adaptable models that can learn from diverse data sources, understand complex traffic scenarios, and react efficiently in real-time, ultimately paving the way for safer and more reliable autonomous vehicles.
 
	\section{Conclusion} 
	The emergence of foundation models has fundamentally reshaped the landscape of artificial intelligence, and their potential for revolutionizing autonomous driving is undeniable. This paper delved into the crux of forging a vision foundation model (VFM) specifically for autonomous driving, highlighting the critical techniques of data generation, pre-training, and adaptation.
	However, the journey towards robust and adaptable autonomous driving perception systems remains challenging. 
	We hope our investigation and platform could boost future research for vision foundation model in safety-critical autonomous driving.


 

        %
        %
        %
        %
        %
        %
        %
        %
        
	{	
	\bibliography{ref}
	}
		
\end{document}